\documentclass[journal]{IEEEtran}
\ifCLASSINFOpdf
\usepackage{hyperref}       
\usepackage{url}            
\usepackage{booktabs}       
\usepackage{amsfonts}       
\usepackage{nicefrac}       
\usepackage{microtype}      
\usepackage{xcolor}         
\usepackage{graphicx}
\usepackage{subfigure}
\usepackage{amsthm,amsmath,amssymb}
\usepackage{mathrsfs}
\usepackage{amssymb}
\usepackage{multirow}
\usepackage{cite}
\definecolor{gg}{RGB}{6,128,67}
\else
\fi

\begin{document}
%
\title{Wavelet Prior Attention Learning in Axial Inpainting Network}

%
%
%

\author{Chenjie~Cao,~
        Chengrong~Wang,~
        Yuntao~Zhang,~
        Yanwei~Fu
\thanks{
Chenjie~Cao, Chengrong~Wang, Yuntao~Zhang, and Yanwei~Fu are with the school of data science of Fudan University, Shanghai 200433, China. (Corresponding author: Yanwei Fu.)}}

%
%

\markboth{Journal of \LaTeX\ Class Files,~Vol.~14, No.~8, August~2015}%
{Shell \MakeLowercase{\textit{et al.}}: Bare Demo of IEEEtran.cls for IEEE Journals}
%



\maketitle

\begin{abstract}
Image inpainting is the task of filling masked or unknown regions of an image with visually realistic contents, which has been remarkably improved by Deep Neural Networks (DNNs) recently.
Essentially, as an inverse problem, the inpainting has the underlying challenges of 
reconstructing semantically coherent results without texture artifacts.
Many previous efforts have been made via exploiting attention mechanisms and prior knowledge, such as edges and semantic segmentation. However, these works are still limited in practice by an avalanche of learnable prior parameters and prohibitive computational burden.
To this end, we propose a novel model -- Wavelet prior attention learning in Axial Inpainting Network (WAIN), whose generator contains the encoder, decoder, as well as two key components of Wavelet image Prior Attention (WPA) and stacked multi-layer Axial-Transformers (ATs). Particularly, the WPA guides the high-level feature aggregation in the multi-scale frequency domain, alleviating the textual artifacts. 
Stacked ATs employ unmasked clues to help model reasonable features along with low-level features of horizontal and vertical axes, improving the semantic coherence.
Extensive quantitative and qualitative experiments on Celeba-HQ and Places2 datasets are conducted to validate that our WAIN can achieve state-of-the-art performance over the competitors. The codes and models will be released.
\end{abstract}


%
\IEEEpeerreviewmaketitle

\section{Introduction}

\IEEEPARstart{I}MAGE inpainting is the task of recovering missing regions for the given corrupted image. As one of the widely used image processing applications, it is still challenging to recover the realistic and semantic contents compatibly with the holistic structural borderlines.
Recently, Deep Neural Networks (DNNs) have achieved great improvements in image inpainting by maximizing the likelihood over the large-scale training data~\cite{guillemot2013image}.
Compared with traditional approaches~\cite{criminisi2003object,guo2017patch,li2017localization,liu2018structure} with hand-crafted feature matchings, DNN based methods can generate visually realistic and semantically consistent results for large missing holes~\cite{iizuka2017globally,yu2018generative,yu2019free,nazeri2019edgeconnect,zhao2021comodgan,sun2022learning,wang2022ft}. 

\begin{figure}
\begin{centering}
\includegraphics[width=0.99\linewidth]{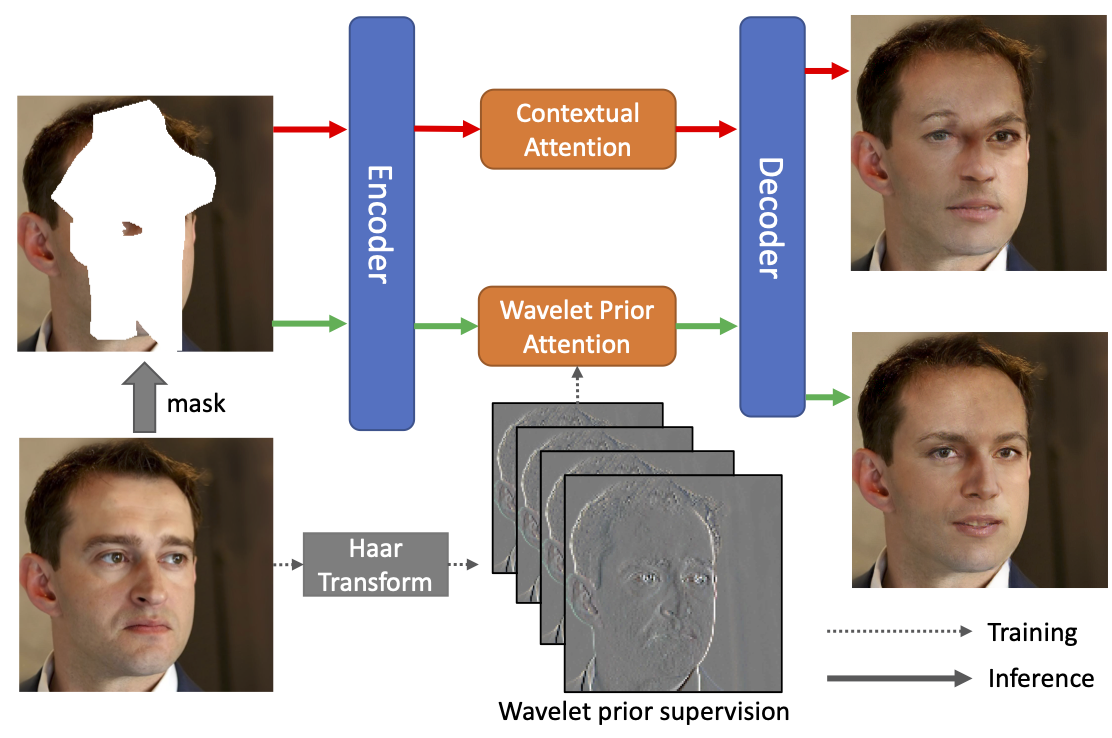} 
\par\end{centering}
\vspace{-0.1in}
\caption{Inpainting comparisons among EdgeConnect (EC)~\protect\cite{nazeri2019edgeconnect}, EC with Gated Convolution (GC)~\protect\cite{yu2019free} and Contextual Attention (CA)~\protect\cite{yu2018generative} (EC+GC+CA), Co-Modulation GAN (Co-Mod)~\protect\cite{zhao2021comodgan}, and our method. Learning semantically reasonable and consistent features for masked regions in (I) and distinguishing confusing textures with different frequencies in (II) are still difficult for existing inpainting methods even with attention schemes. Best viewed with zoom-in.}
\vspace{-0.1in}
\label{figure_teaser}
\end{figure}

Fundamentally as one challenging inverse problems~\cite{starck2009overview},  the ability to learn good image inpainting is still limited in practice by filling the missing image regions being consistent and coherent with the known object and rest of image regions. The underlying challenges can thus be broadly characterized as the \textit{semantic incoherence} and  \textit{texture artifacts} of the generated image contents.

\textit{Inpainted result is semantically incompatible to unmasked regions}. Since many plausible solutions can be achieved for the given context, the synthesis usually suffers from incompatible and unnatural results compared with uncorrupted regions. \emph{e.g.}, directions (eye gaze) of generated right eye and unmasked left eye are inconsistent in Figure~\ref{figure_teaser} (I). 
Additionally, this problem also negatively affects the generation of some structural details in scenes inpainting.

\textit{The generated content has many texture artifacts}. Some image pixels enjoy similar colors and textures, which makes DNNs tend to generate fuzzy hallucinations or ambiguous texture artifacts or borderlines as visualized in Figure~\ref{figure_teaser} (II). These artifacts are caused by synthesizing the content from different frequency domains.

One strategy of addressing the challenge of semantic incoherence is via exploring better feature representations of the masked areas, as in previous efforts\cite{iizuka2017globally,yu2019free,nazeri2019edgeconnect,yang2020learning,xie2019image,zhu2021image}. 
For example, dilated convolutions~\cite{yu2015multi} are leveraged in~\cite{iizuka2017globally} to expand the respective fields for masked regions, which is also adopted in many works~\cite{yu2019free,nazeri2019edgeconnect,yang2020learning}.
Furthermore, Partial Convolutions~\cite{Liu_2018_ECCV} and Gated Convolutions (GC)~\cite{yu2019free} are proposed to manually and adaptively combine the masked and unmasked features.
Unfortunately, previous approaches fail to capture the holistic semantics of unmasked regions, and make the generated contents coherence. Such as EdgeConnect (EC)~\cite{nazeri2019edgeconnect}, EC with GC~\cite{nazeri2019edgeconnect,yu2019free} are unable to explicitly model the holistic semantics such as eye gaze.
Critically, such semantic incoherence cannot even be solved by StyleGAN2~\cite{karras2020analyzing} based Co-Modulation GAN (Co-Mod)~\cite{zhao2021comodgan} learned on large-scale data as shown in Figure~\ref{figure_teaser} (I)-(d). 
This motivates our new exploration of learning the more compatible and meaningful features from the masked and unmasked regions.

To overcome the texture artifacts, many previous works resort to model image priors for image inpainting, such as structure-based priors with edges~\cite{nazeri2019edgeconnect,lin2020foreground}, edge smoothing structures~\cite{liu2020rethinking}, and facial landmarks~\cite{yang2019lafin}, image gradient-based priors~\cite{yang2020learning}, and semantic segmentation priors~\cite{liao2020guidance,liao2021image}. Unfortunately, the priors modeled in these works are not only strongly dependent on the unmasked image regions, but also may potentially demand significant computational costs for multi-stage models.  
Besides, attention mechanisms~\cite{vaswani2017attention} have been introduced to image inpainting recently~\cite{yu2018generative,yu2019free,zeng2019learning,yi2020contextual,wan2021high,yu2021diverse,wu2021deep}. Despite its intuitiveness and effectiveness, most of previous attention based inpainting methods \emph{e.g.}, Contextual Attention (CA)~\cite{yu2018generative} still suffer from the problems of semantic incoherence and texture artifacts, as shown in Figure~\ref{figure_teaser} (c). And the standard attention modules~\cite{wan2021high,yu2021diverse,esser2021imagebart,wang2022ft} also fail to tackle these detailed issues. 
Even worse, it generally requires prohibitive computational cost and memory footprint to compute the attention.  It is thus necessary to improve both efficiency and effectiveness of attention for image inpainting. 

In this paper, we propose solving these challenges by using 
attention mechanism to aggregate and encode the feature representations. Specifically, we present the Wavelet image Prior Attention for feature aggregation. This component models the template-based image prior by wavelet-guided contextual attention, and thus better learn high-level prior of the whole image in low-resolution CNN features. Essentially, the WPA employs the Haar Transform (HT) to supervise the learning of the feature attention map, and significantly reduce the texture artifacts. Furthermore, the template HT-based wavelet priors help efficiently recover the multi-scale frequent information with very limited extra training computation of attention. On the other hand, the feature encoding exploits multi-layer stacked Axial-Transformers~\cite{ho2019axial} to encode low-level features with high-resolution. This can help improve the semantic coherence of inpainted regions. Compared with fully convolutional encoders, learning stacked attentions can utilize more clues and details of unmasked regions to reduce the semantic contradictions.

Formally, this paper proposes a novel model -- Wavelet prior attention learning in Axial Inpainting Network (WAIN). It has the key components of learning Wavelet Prior Attention (WPA) and stacking multi-layer Axial-Transformers (ATs). 
For the WPA, we firstly get coarse-to-fine wavelet frequencies from masked images with HT. 
Then, these corrupted wavelets are aggregated and restored by the cosine similarities from CA.  
Afterward, the inverse HT is utilized to upscale the low-resolution images with aggregated wavelets. 
To optimize the attention with wavelet priors, these multi-scale aggregated wavelets and related up-scaled images are supervised with the ground-truth ones. Thus, the multi-scale aggregation is leveraged to strengthen the frequency relation of CA in high-level features. 
Note that the proposed wavelet prior attention is a lightweight module since no extra parameters are introduced for aggregating the supervision schema. 
Besides, dilated convolutions for low-level feature encoding are replaced by stacked ATs. Learning masked features with vertical and horizontal correlations in ATs facilitates generating semantically coherent results for WAIN. Moreover, AT enjoys more efficient memory and computation costs compared with the standard attention, which allows us to stack more AT layers in low-level features for superior performances.

Quantitative and qualitative image inpainting experiments are conducted on Celeba-HQ~\cite{CelebAMask-HQ} and Places2~\cite{zhou2017places} dataset to validate the efficacy of our model over the competitors. Furthermore, adequate ablation studies are discussed to illustrate the effectiveness of the proposed components.
Contributions of this paper can be highlighted as follows, 
1) We propose a novel wavelet prior attention module, which utilizes the template-based function --wavelet to model the image prior. Such a module has no learnable parameters and few extra computations to model the wavelet prior of images. 
2) The multi-scale strategy has been adopted to further combine wavelet prior with an attention mechanism, to capture more details and correlations from various frequencies without heavy computations. 
3) We present the stacked Axial-Transformer blocks to encode features of the masked regions. Such blocks can effectively learn the more meaningful features based on unmasked clues with arbitrary distances from the corrupted image.

\section{Related Work}

\subsection{Feature Encoding for Image Inpainting}
Since DNNs are introduced to image inpainting, many works have tried to learn semantically consistent features for missing regions with learning-based models. Pathak \textit{et. al.}~\cite{pathak2016context} firstly leverage an encoder-decoder based model trained with adversarial loss to generate missing pixels. Then, Iizuka \emph{et. al.}~\cite{iizuka2017globally} propose to use dilated convolutions to expand the respective fields for masked regions. They also train two discriminators for local masked regions and global inpainted images, respectively. Furthermore, partial convolutions~\cite{Liu_2018_ECCV} mask can re-normalize the convolutions conditioned on valid regions for irregular masks. The gated convolutions~\cite{yu2019free} propose a simple way to encode inpainting features with the gating mechanism learned by the network adaptively. Xie \emph{et al.}~\cite{xie2019image} propose a learnable attention map to learn feature re-normalization and mask-updating. Moreover, the works of ~\cite{zhu2021image} and \cite{yan2021agg} both use the multi-scale strategy for point-wise normalization and dilated convolutions for masked feature encoding individually. Despite many efforts have been taken, it is still very challenging and yet crucial to simultaneously learning meaningful masked features with arbitrary missing regions and generating visually realistic results. Different from previous efforts, we stack Axial-Transformers for the masked feature learning with the efficient axial-based attention mechanism. The novel transformer layers can efficiently and effectively encode the compatible masked features with the clues in the valid regions.

\subsection{Image Inpainting by Modeling Priors}
Various types of image priors have been exploited to facilitate image inpainting. Liao \emph{et al.}~\cite{liao2020guidance,liao2021image} employ semantic segmentation priors to ensure semantic consistency, while Yang \emph{et al.}~\cite{yang2019lafin} utilize facial landmarks to generate the plausible appearance. The works of~\cite{nazeri2019edgeconnect,lin2020foreground} leverage edge sketches of uncorrupted areas as the prior to guide the inference of missing pixels. This type of prior is further generalized in 
Liu \emph{et al.}~\cite{liu2020rethinking} which models the edge-preserved smooth image to represent the global structure of image scenes. Furthermore, in~\cite{yang2020learning}, this work uses Sobel filters to extract gradient maps to supervise intermediate features in decoders. 
Besides, Wu \emph{et al.} leverage Local Binary Pattern (LBP) maps to strengthen the inpainting~\cite{wu2021deep} with more information compared with canny edge.
In~\cite{yu2021wavefill}, Yu \emph{et al.} optimize the low and high-frequency images with $l_1$ and adversarial loss separately, and use attention to fuse features with different frequencies.
The work that is closest to ours is WaveFill~\cite{yu2021wavefill}, which recovers low-frequency and high-frequency separately with a multiple optimization stages and two discriminators, resulting in heavy computational cost. In contrast, we re-purpose the wavelet prior, and formulate the wavelet prior attention module to model various frequencies to guide our inpainting network efficiently.


      
\subsection{Attention-based Image Inpainting}
Inpainting the missing region in an image requires a global understanding of its texture and structure, for which the attention mechanism can demonstrate excellent performance by modeling long-range correlations. Yu \emph{et al.}~\cite{yu2018generative} firstly proposed contextual attention to collect similar feature patches from the background to fill in masked region. Then Zeng \emph{et al.}~\cite{zeng2019learning} transfer attention relation at multiple abstraction levels, while the work in~\cite{yi2020contextual} further generalizes this idea by extending the attention transfer module to residual map and generate high-resolution results. Moreover, Zhou \emph{et al.}~\cite{zhou2020learning} propose the attention maps of unmasked images to supervise the masked ones. Furthermore, reference~\cite{liao2021image} propagates attention within the same semantic class to prevent blurry boundaries and semantic confusion. Besides, some works~\cite{wan2021high,yu2021diverse} leverage standard attention modules to tackle low-resolution restoration for the two-stage inpainting architecture.
Overall, existing strategies of supervising attention bring significant computation costs in the attention transfer on multi-level feature maps across different semantics or multi-stage inpainting models. In contrast, our model utilizes the template-based wavelet to supervise the attention with image priors. It thus keeps a lightweight architecture without significant extra computational cost.

\section{Methodology}


\begin{figure*}
\begin{centering}
\includegraphics[width=0.9\linewidth]{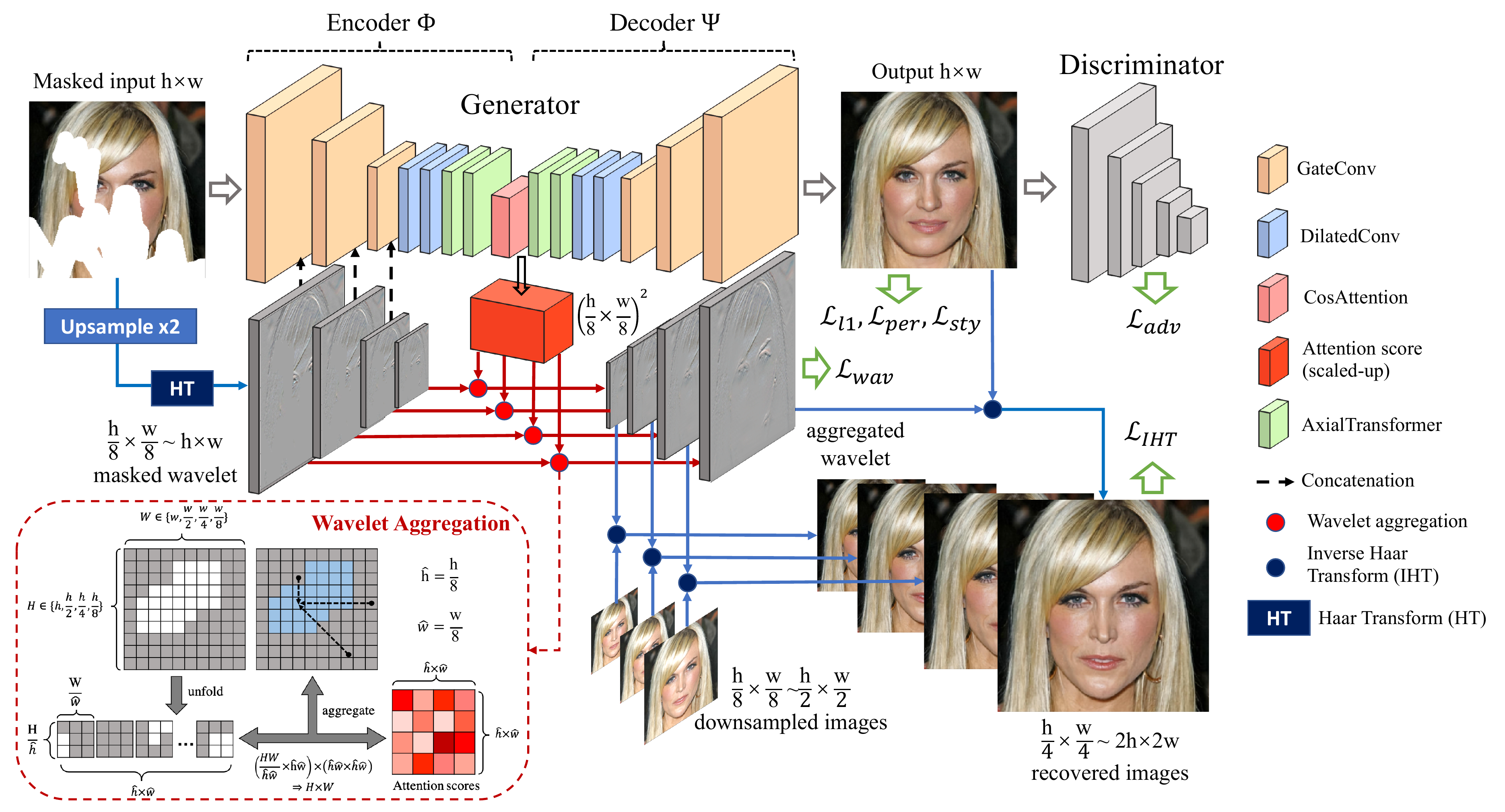} 
\par\end{centering}
\vspace{-0.1in}
\caption{Overview of WAIN. The masked input is firstly upsampled before the HT. Then, multi-scale wavelets are extracted by the HT and then concatenated to related encoder features. WPA between the encoder and the decoder directly aggregates the wavelet maps of different scales. Aggregated wavelets further recover higher-resolution images with low-resolution ground truths by IHT. Dashed box in the bottom-left shows details about the wavelet aggregation. \label{figure_overview} }
\vspace{-0.1in}
\end{figure*}

\subsection{Model Overview}

Our WAIN is overviewed in Figure~\ref{figure_overview}. WAIN is an adversarial training system consisting of a generator and a discriminator. The generator is composed of an encoder $\Phi$, a decoder $\Psi$, and the WPA (Sec.~\ref{WPA}). For better generalization on irregular masked regions, we use GCs~\cite{yu2019free} for convolution layers in the generator with stride 2 as in~\cite{cao2021learning}. 
Then, the intermediate layers of $\Phi$ and $\Psi$ are composed of 4 dilated convolutions and 4 axial-transformers (Sec.~\ref{AT}). And a wavelet prior based attention module is implemented in the generator for unfolding and aggregating both features and wavelets with patches. To introduce the wavelet prior, we upsample the masked input images and then decouple them into coarse-to-fine wavelet features with HT. Therefore, input images for WAIN have the same resolution compared with the baseline and other competitors. Besides, we take the SN-PathGAN~\cite{yu2019free} as the discriminator.

\subsection{Cosine Attention Mechanism}

The proposed WPA is built upon the contextual attention (CA)~\cite{yu2018generative}, which can also be formulated as cosine attention. 
Typically, the classical  attention~\cite{vaswani2017attention} is as
\begin{equation}
\textrm{Attention}(\mathbf{Q},\mathbf{K},\mathbf{V})=\mathrm{softmax}(\frac{\mathbf{Q}\mathbf{K}^T}{\sqrt{d}}-inf\cdot \mathbf{M})\mathbf{V},
\label{4}
\end{equation}
where $d$ is the input feature channel. Binary matrix  $\mathbf{M}$ makes some inner product values of $\mathbf{Q}\mathbf{K}^T$ in masked regions to negative infinity $-inf$.
The $\mathbf{Q}, \mathbf{K}, \mathbf{V}$ are computed from the same input projected with different parameter matrices. 
Different from the above attention in Eq.~(\ref{4}), CA removes the parametric projections for key, query, and value. Specifically, we compute the cosine similarity by normalizing the input features to ensure maximum attention scores on the diagonal for unmasked regions. Given the masked region $\mathbf{M}$, CA unfolds the input features into uncorrputed feature patches $\mathbf{F}_u$ from the unmasked regions $\mathbf{F}_{\mathbf{1}-\mathbf{M}}$ and the corrupted feature patches $\mathbf{F}_m$ from masked regions $\mathbf{F}_\mathbf{M}$. Then the normalized cosine attention score can be got as
\begin{equation}
\begin{split}
\mathrm{cos}_{u,m}&=\left<\frac{\mathbf{F}_u}{||\mathbf{F}_u||},\frac{\mathbf{F}_m}{||\mathbf{F}_m||}\right>\\
\mathbf{R}_{u,m}&=\mathrm{softmax}_{u}(\lambda \mathrm{cos}_{u,m}),
\end{split}
\label{5}
\end{equation}
where $\mathrm{softmax}_{u}$ indicates the softmax operation for unmasked patches, and $\lambda=10$ is the constant `temperature' to enlarge the variance of attention scores. Subsequently, the output feature patches $\mathbf{F}'_m$ are aggregated from the unmasked feature patches $\mathbf{F}_u$ according to the attention score 
\begin{equation}
\mathbf{F}'_m=\sum_{u\in \mathbf{1}-\mathbf{M}} \mathbf{R}_{u,m}\mathbf{F}_u.
\label{6}
\end{equation}
In practice, Eq.~(\ref{5}) and Eq.~(\ref{6}) work in parallel for every sample in~\cite{yu2018generative}. So the attention relation $\mathbf{R}\in \mathbb{R}^{\hat{h}\hat{w}\times\hat{h}\hat{w}}$ is a matrix, where $\hat{h}, \hat{w}$ are the height and width of patches unfolded from the feature maps. And $\mathbf{R}_{i,j}, (i\leq \hat{h}\hat{w}, j\leq \hat{h}\hat{w})$ means attention score between patch $i$ and patch $j$. Empirically, we show that normalization helps to get better results in image inpainting as in Sec.~\ref{Quantitative Ablation of Different Attention Mechanisms}. 

\subsection{Wavelet Prior Attention \label{WPA}}
The aggregation of Eq.~(\ref{6}) is not enough to produce satisfying inpainting results as shown in Figure~\ref{figure_teaser}~(c). Essentially, as cosine similarities are completely computed from encoded image features, it will prefer the local consistency of color and texture and ignore the correlations between low and high frequency information. However, these correlations are desirable to reduce the texture artifacts. 

To this end, the wavelet prior based attention is proposed and shown in Figure~\ref{figure_overview}. For the given masked image $\mathbf{I}_m\in \mathbb{R}^{h\times w\times c}$, where $c=3$ with RGB, we use the discrete wavelet of Haar Transform (HT) with fixed decomposition filters to decompose it into a downscaled representation $\mathbf{I}^{HT}_m\in \mathbb{R}^{\frac{h}{2}\times \frac{w}{2}\times c}$ and high-frequency features of three directions $\mathbf{H}^{HT}_m\in \mathbb{R}^{\frac{h}{2}\times \frac{w}{2}\times 3c}$ ~\cite{huang2017wavelet,xiao2020invertible}. 
Note that the input and output image sizes are both $h\times w$, so wavelets with $h\times w$ cannot be obtained unless inputs are enlarged to $2h\times 2w$. Rather than directly using higher-resolution inputs, we adopt a strategy of simply 
 upsampling input images to $\mathbf{I}^{(0)}_m\in \mathbb{R}^{2h\times 2w\times 3}$ with bilinear interpolation. Empirically, this has better performance as discussed in Sec.~\ref{Quantitative Ablation of WPA}.
 
HT is repeated for $L=4$ times to get the multi-scale high-frequency features $\mathbf{H}^{(l)}_m\in \mathbb{R}^{\frac{h}{2^{l-1}}\times\frac{w}{2^{l-1}}\times 3c}$,
\begin{equation}
\mathbf{I}^{(l)}_m, \mathbf{H}^{(l)}_m=\mathrm{HT}(\mathbf{I}_m^{(l-1)}), (1<l\leq L).
\label{7}
\end{equation}
We still use $\mathbf{I}_m\in \mathbb{R}^{h\times w\times 3}$ as the input image to the inpainting model. 
Then, these coarse-to-fine masked $\mathbf{H}^{(l)}_m$ are aggregated by the attention relation matrix $\mathbf{R}$ got from Eq. (\ref{5}) as shown in the sub-figure at bottom-left of Figure~\ref{figure_overview}. Specifically, to aggregate multi-scale features with the same attention relation, we firstly unfold the wavelet feature $\mathbf{H}_m^{(l)}$ according to the shape of attention relation $\mathbf{R}\in \mathbb{R}^{\hat{h}\hat{w}\times \hat{h}\hat{w}}$. The unfolded numbers of rows and columns are $\hat{h}=\frac{h}{8},\hat{w}=\frac{w}{8}$ correspondingly. So the unfolded wavelet features $\mathbf{H}^{(l)}_{i}$ can be formulated as 
\begin{equation}
\begin{split}
\negthickspace
\negthickspace
\mathbf{H}^{(l)}_{1}, \mathbf{H}^{(l)}_{2}, ...,\mathbf{H}^{(l)}_{\hat{h}\hat{w}}&=\mathrm{unfold}(\mathbf{H}^{(l)}, \mathrm{row}=\hat{h}, \mathrm{col}=\hat{w}),\\
\mathbf{H}^{(l)}_{i}\in \mathbb{R}&^{\frac{h}{2^{l-1}\hat{h}}\times\frac{w}{2^{l-1}\hat{w}}\times 3c}, (1\leq i\leq \hat{h}\hat{w}).
\end{split}
\label{8}
\end{equation}
Then the unfolded wavelet patches are aggregated by the weighted summation of the attention relation $\mathbf{R}$ as
\begin{equation}
\mathbf{H}'^{(l)}_j=\sum_i^{\hat{h}\hat{w}}\mathbf{R}_{i,j}\mathbf{H}^{(l)}_i, (i\leq \hat{h}\hat{w}, j\leq \hat{h}\hat{w}),
\label{9}
\end{equation}
where $i,j$ are the indices of $\mathbf{R}\in \mathbb{R}^{\hat{h}\hat{w}\times \hat{h}\hat{w}}$. 

For a faster batch calculation, we re-implement CA in group convolutions for both cosine similarity and feature aggregation. We use the balanced $l_1$-loss to supervise the aggregated wavelet features with the ground-truth wavelet features $\mathbf{\hat{H}}^{(l)}$ as follows,
\begin{small}
\begin{equation}
\begin{split}
\mathcal{L}_{wav} & =\mathbb{E}\left[\sum_{l=1}^L\frac{1}{N_{\mathbf{M}}}\lVert\mathbf{M}\odot(\mathbf{\hat{H}}^{(l)}-\mathbf{H}'^{(l)})\rVert_{1}\right]\\
 & +\mathbb{E}\left[\sum_{l=1}^L\frac{1}{N_{(\mathbf{1}-\mathbf{M})}}\lVert(\mathbf{1}-\mathbf{M})\odot(\mathbf{\hat{H}}^{(l)}-\mathbf{H}'^{(l)})\rVert_{1}\right],
\end{split}
\label{10}
\end{equation}
\end{small}
\noindent where $\mathbf{M}$ means the mask map, and $N_\mathbf{M}, N_{(\mathbf{1}-\mathbf{M})}$ indicate pixel sums of masked and unmasked regions for the normalization. Meanwhile, the Inverse Haar Transform (IHT), which double-upsamples the input low-resolution images with high-frequency wavelet features, is used to get the balanced $l_1$-loss of ground-truth multi-scale images $\mathbf{\hat{I}}^{(l)}$ to further supervise the CA as follows
\begin{equation}
\begin{split}
\mathbf{\tilde{I}}^{(l)}=\mathrm{IHT}&(\mathbf{\hat{I}}^{(l+1)}, \mathbf{H}'^{(l+1)}), (1\leq l\leq L-1),\\
\mathbf{\tilde{I}}^{(l)}&=\mathrm{IHT}(\mathbf{\tilde{I}}, \mathbf{H}'^{(l+1)}), (l=0),
\label{11}
\end{split}
\end{equation}
\begin{equation}
\begin{split}
\mathcal{L}_{IHT} & =\mathbb{E}\left[\sum_{l=0}^{L-1}\frac{1}{N_{\mathbf{M}}}\lVert\mathbf{M}\odot(\mathbf{\hat{I}}^{(l)}-\mathbf{\tilde{I}}^{(l)})\rVert_{1}\right]\\
 & +\mathbb{E}\left[\sum_{l=0}^{L-1}\frac{1}{N_{(\mathbf{1}-\mathbf{M})}}\lVert(\mathbf{1}-\mathbf{M})\odot(\mathbf{\hat{I}}^{(l)}-\mathbf{\tilde{I}}^{(l)})\rVert_{1}\right],
\end{split}
\label{12}
\end{equation}
\noindent where $\mathbf{\tilde{I}}\in \mathbb{R}^{h\times w\times3}$ indicates the inpainted output image. And $\mathbf{\tilde{I}}$ is used as the low-frequency image for $l=0$ in $\mathcal{L}_{IHT}$ instead of the ground truth one. Besides, the ground truth $\mathbf{\hat{I}}^{(0)}$ is resized from the uncorrupt input $\mathbf{I}\in \mathbb{R}^{h\times w\times3}$ without leveraging any additional high-resolution information.



We summarize the advantages of our WPA module. The proposed wavelet prior attention is much more efficient without feature encoding and concatenating. In particular,
it encourages the model to learn the latent correlation between masked $\mathbf{F}_{\mathbf{M}}$ and unmasked features $\mathbf{F}_{\mathbf{1}-\mathbf{M}}$ to reconstruct the distribution of the missing high-frequency feature $\mathbf{H}_{\mathbf{M}}$ through the unmasked one $\mathbf{H}_{\mathbf{1}-\mathbf{M}}$.
This process has no trainable parameters. Furthermore, WPA serves as the auxiliary supervision which improves the attention learning by making the inpainting model WAIN distinguish confusing visual regions with only image features during the inference (Sec.~\ref{Qualitative Ablation Study of WPA}).

\subsection{Axial Transformer for Inpainting}
\label{AT}

\begin{figure}
\begin{centering}
\includegraphics[width=0.9\linewidth]{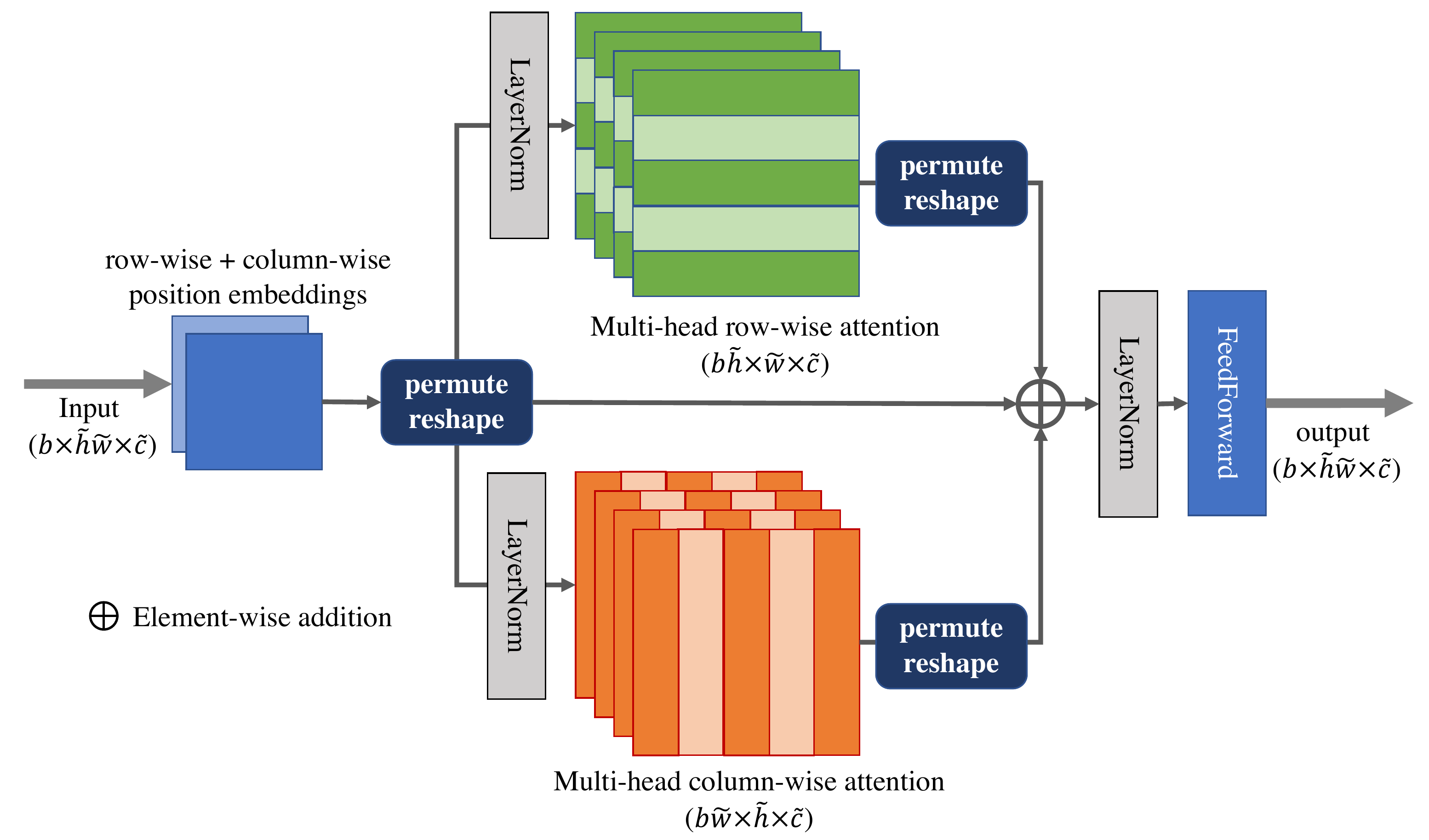} 
\par\end{centering}
\vspace{-0.1in}
\caption{Illustration of the axial-transformer block used in WAIN. Axial-position embeddings are only added before the first axial-transformer block. The `permute reshape' means adjusting the tensor shape for row-wise and column-wise attention calculations respectively.}
\label{figure_AT} 
\vspace{-0.1in}
\end{figure}

Dilated convolutions~\cite{iizuka2017globally,nazeri2019edgeconnect,yu2019free,yan2021agg} (DCs) have been widely used by deep inpainting models in expanding the respective fields for masked regions. However, DCs are still insufficient to learn visually consistent and semantically reasonable masked features for inpainting.
The key challenge lies in the incapability of extracting meaningful features of arbitrary masked regions even by dilated kernels.
It may potentially break the semantic coherence with unnatural inpainted contents. For example, there are asymmetrical eyes and glasses as illustrated in Sec.~\ref{Qualitative Ablation Study of AT}. This can not be solved with simple attention mechanisms for high-level features.

We resort to Axial-Transformer (AT) for our inpainting model, which is firstly proposed in~\cite{ho2019axial}, and has been utilized in some vision autoregressive models~\cite{wang2020axial,kumar2021colorization}.
AT calculates the attention in rows and columns respectively, and it can tame the $O(n^2)$ complexity in computing the standard self-attention. 
In both the encoder and decoder of our WAIN, we replace DCs with ATs and thus better encode the low-level features. Compared with DCs, ATs can achieve larger respective fields and model more reliable dependencies and correlations between masked and unmasked regions. Moreover, the low-level features (64$\times$64) learned by ATs enable better visual clues in the unmasked regions and facilitate compatible reconstruction. 

For the input batch features $\mathbf{X\in\mathbb{R}}^{b\times\tilde{h}\tilde{w}\times\tilde{c}}$ in immediate layers of the generator, $\tilde{h},\tilde{w}=\frac{h}{4},\frac{w}{4}$; $b$ is the batch size; and $\tilde{c}$ is the channels of the feature. We add the row-wise and column-wise position embeddings before the first AT module,
\begin{equation}
\mathbf{X}'_{i,j}=\mathbf{X}_{i,j}+\mathrm{Embedding}_{row}(i)+\mathrm{Embedding}_{col}(j),
\label{AT1}
\end{equation}
where $i\in\tilde{h},j\in\tilde{w}$ are indices of row and column respectively. 
Positional priors are beneficial to recovering data with relatively stable distribution, such as human faces. 
A simple way to implement ATs is to split the feature $\mathbf{X}'$ into row-wise and column-wise features. So we permute and reshape the input tensor to $\mathbf{X}_{row}\in\mathbb{R}^{b\tilde{h}\times\tilde{w}\times\tilde{c}}$ and $\mathbf{X}_{col}\in\mathbb{R}^{b\tilde{w}\times\tilde{h}\times\tilde{c}}$ for the multi-head row-wise and column-wise attentions respectively as shown in Figure~\ref{figure_AT}. We use the standard multi-head attention~\cite{vaswani2017attention} to encode these features. Then, the row-wise and column-wise features are reshaped back and summed up with the input features. We also add a FeedForward layer consisted of two linear layers and a GELU activation~\cite{vaswani2017attention} after the attention. 

Note that, differs from ATs in~\cite{ho2019axial}, we simply leverage ATs without the autoregressive training and causal masks~\cite{ho2019axial,kumar2021colorization} as in Figure~\ref{figure_AT}.
Thus we do not use any mask in ATs. AT is a substitute for DC to better encode the masked features; so it also needs to learn encoding masked features themselves. Furthermore, if any rows or columns of an image are all masked, there will be some numerical problems in computing the attention scores with masked ATs.

AT requires significantly fewer computations, because it only aggregates the features along a single row or column. Compared with the standard attention~\cite{vaswani2017attention} of $O(b(\tilde{h}\tilde{w})^2)$, the computation of AT is $O(b(\tilde{h}\tilde{w}^2+\tilde{w}\tilde{h}^2))$.
Stacked ATs with larger input feature maps are still good at modeling the superior long-range spatial dependencies of masked and unmasked regions. AT can outperform standard attention due to its efficiency to tackle low-level features. For example, with the same memory footprint, our model can train a generator with 4 AT blocks in $64\times64$ features; in contrast, the standard attention can be only trained with $32\times32$ features.
More critically, our ATs are complementary to the WPA module, as empirically evaluated in Sec.~\ref{Quantitative Ablation Study of AT} and \ref{Qualitative Ablation Study of AT}.



\subsection{Loss Functions}

We train the proposed WAIN inpainting model with the adversarial loss~\cite{goodfellow2014generative} as in Figure~\ref{figure_overview}. Given the output image $\mathbf{\tilde{I}}\in \mathbb{R}^{h\times w\times 3}$ from the generator $G$, the SN-PathGAN~\cite{yu2019free} based discriminator is implemented as $D$, and the adversarial loss for discriminator and generator is
\begin{equation}
\vspace{-0.1in}
\begin{split}
\mathcal{L}_{D}=-\mathbb{E}_{\mathbf{\hat{I}}}[\log &D(\mathbf{\hat{I}})]-\mathbb{E}_{\mathbf{\tilde{I}}}[1-\log D(\mathbf{\tilde{I}})],\\
\mathcal{L}_{adv}&=-\mathbb{E}_{\mathbf{\tilde{I}}}[\log D(\mathbf{\tilde{I}})],
\end{split}
\label{16}
\vspace{-0.1in}
\end{equation}
where $\mathbf{\hat{I}}$ means the ground truth image. For the stable adversarial training, we add spectral normalization~\cite{miyato2018spectral} to both the discriminator and the generator~\cite{odena2018generator}. Note that we use the vanilla adversarial training loss~\cite{goodfellow2014generative}, as other GAN losses can not achieve significant improvements in our model. We also employ the reconstruction losses  with $l_1$-loss $\mathcal{L}_{l_1}$ and VGG-19 based perceptual loss $\mathcal{L}_{per}$~\cite{johnson2016perceptual} with the balanced form between masked and unmasked regions 
\begin{equation}
\vspace{-0.1in}
\begin{split}
\mathcal{L}_{l_1} & =\mathbb{E}\left[\frac{\lVert\mathbf{M}\odot(\mathbf{\hat{I}}-\mathbf{\tilde{I}})\rVert_{1}}{N_{\mathbf{M}}}\right] +\mathbb{E}\left[\frac{\lVert(\mathbf{1}-\mathbf{M})\odot(\mathbf{\hat{I}}-\mathbf{\tilde{I}})\rVert_{1}}{N_{(\mathbf{1}-\mathbf{M})}}\right],\\
\mathcal{L}_{per} & =\mathbb{E}\left[\sum_{k=1}^{K}\frac{\lVert\mathbf{M}\odot(\Phi_k^{\mathbf{\hat{I}}}-\Phi_k^{\mathbf{\tilde{I}}})\rVert_{1}}{N_{\mathbf{M}_k}}\right] \\
&+\mathbb{E}\left[\sum_{k=1}^{K}\frac{\lVert(\mathbf{1}-\mathbf{M})\odot(\Phi_k^\mathbf{{\hat{I}}}-\Phi_k^{\mathbf{\tilde{I}}})\rVert_{1}}{N_{(\mathbf{1}-\mathbf{M}_k)}}\right],\\
\end{split}
\label{eq:l1_and_per}
\vspace{-0.1in}
\end{equation}
where $\odot$ means element-wise multiplication; $\mathbf{M}$ means binary mask map; $\Phi_k^{\mathbf{\hat{I}}}$, $\Phi_k^{\mathbf{\tilde{I}}}$ represent VGG-19 features from layer $k\in K$ with fake and real inputs respectively; $N_{\mathbf{M}}$, $N_{(\mathbf{1}-\mathbf{M})}$, $N_{\mathbf{M}_k}$, $N_{(\mathbf{1}-\mathbf{M}_k)}$ indicate normalization coefficients for different mask maps. 
Moreover, style loss $\mathcal{L}_{sty}$~\cite{gatys2016image} between the output images and ground-truth images is also incorporated as 
\begin{equation}
\mathcal{L}_{sty} = \mathbb{E}\left[\lVert G^{\Phi_k}(\mathbf{\hat{I}})-G^{\Phi_k}(\mathbf{\tilde{I}})\rVert_{1}\right],
\label{eq:style_loss}
\end{equation}
where $G^{\Phi_k}$ means the Gram matrix got from VGG features of layer $k$.
Therefore, the total generative loss of our model can be summarized as
\begin{equation}
\begin{split}
\mathcal{L}_{G}=\lambda_{adv}\mathcal{L}_{adv}+\mathcal{L}_{l_1}+\lambda_{per}\mathcal{L}_{per}+\lambda_{sty}\mathcal{L}_{sty}\\
+\mathcal{L}_{wav}+\lambda_{IHT}\mathcal{L}_{IHT},
\end{split}
\label{17}
\end{equation}
where we empirically set $\lambda_{adv}=0.1, \lambda_{per}=0.1, \lambda_{sty}=250$, and $\lambda_{IHT}=0.5$.

\section{Experiments}

\subsection{Datasets and Mask Settings} 
The proposed approach is evaluated on two various datasets: Celeba-HQ~\cite{CelebAMask-HQ} and Places2~\cite{zhou2017places}. Celeba-HQ contains 30k high-quality human face pictures from celebrities. We select 1k of them as validation images, while the rest are used as training ones. For Places2 with various scenes, we chose 10 categories randomly with 50k training images and 1k validation images, which are ranged from urban to natural scenes.
To train the inpainting model, we randomly take half images with irregular masks~\cite{yu2019free}, and others are corrupted with semantic segmentation masks~\cite{yi2020contextual,zeng2020high}, respectively. 
This masking strategy is used to simulate the object removal mask from the human.  
For the segmentation masks, we collect 130k semantic segmentation masks from COCO dataset~\cite{lin2014microsoft} with mask rate in $[10\%, 40\%]$ of the whole image. 


\subsection{Training Settings}
The proposed method is implemented with PyTorch. 
The original input and output size of the inpainting model are both $256\times256$ as many other inpainting methods. And the input size to Haar transform is upsampled to $512\times512$ from $256\times256$.
We train the model in 240k steps in Celeba-HQ and 1000k steps in Places2 in Adam optimizer~\cite{Adam2015} of $\beta_1=0$ and $\beta_2=0.9$. The initial learning rates are $2e-4$ and $2e-5$ for the generator and discriminator respectively. Then, the learning rate is decayed with 0.5 for $1/5$ of the total steps. Our model is trained in Pytorch v1.3.1, and costs about 1 day to train in Celeba-HQ and 5 days to train in Places2 with a single NVIDIA(R) Tesla(R) V100 16GB GPU.

\subsection{Compared Methods} 
\label{sec:Compared Methods}
We compare the proposed approach with other various state-of-the-art methods. These methods include Gate Convolution (GC)~\cite{yu2019free}, Edge Connect (EC)~\cite{nazeri2019edgeconnect}, Contextual Residual Aggregation (HiFill)~\cite{yi2020contextual}, Multi-Task Structural learning (MTS)~\cite{yang2020learning}, Co-Modulation GAN (Co-Mod)~\cite{zhao2021comodgan}, and WaveFill~\cite{yu2021wavefill}.
For these relatively early methods (GC, EC, HiFill, MTS), we retrain them with our data splits and training settings. For Co-Mod and WaveFill, we use the officially released model weights for face and scene images.
All inpainted results are combined with the unmasked regions of the original images and resized into $256\times256$.

\subsection{Quantitative Comparisons}

\begin{table*}
\center
\vspace{-0.1in}
\caption{Quantitative evaluation results on Celeba-HQ and Places2 with different mask ratios.}
\vspace{-0.1in}
\label{table_quantitative}
\renewcommand\tabcolsep{3.5pt}
\begin{tabular}{c|c|ccccccc|ccccccc}
\hline 
 &  & \multicolumn{7}{c|}{Celeba-HQ} & \multicolumn{7}{c}{Places2}\tabularnewline
\hline 
\hline 
 & Mask & GC & EC & HiFill & MTS & Co-Mod & WaveFill & Ours & GC & EC & HiFill & MTS & Co-Mod & WaveFill & Ours\tabularnewline
\hline 
\multirow{5}{*}{PSNR$\uparrow$} & 10\textasciitilde 20\% & 29.54 & 30.63 & 29.23 & 29.37 & 30.07 & 31.14 & \textbf{31.23} & 27.04 & 29.02 & 28.31 & 28.10 & 27.86 & 28.70 & \textbf{29.44}\tabularnewline
 & 20\textasciitilde 30\% & 26.48 & 27.72 & 26.41 & 26.94 & 27.11 & 28.40 & \textbf{28.45} & 23.69 & 25.98 & 25.35 & 25.66 & 24.70 & 25.88 & \textbf{26.24}\tabularnewline
 & 30\textasciitilde 40\% & 24.42 & 25.75 & 24.53 & 25.12 & 25.13 & 26.42 & \textbf{26.45} & 21.64 & 23.96 & 23.48 & 23.94 & 22.65 & 23.78 & \textbf{24.20}\tabularnewline
 & 40\textasciitilde 50\% & 22.85 & 24.13 & 23.06 & 23.72 & 23.65 & 24.85 & \textbf{24.89} & 20.05 & 22.42 & 22.03 & 22.54 & 21.14 & 21.54 & \textbf{22.63}\tabularnewline
 & Mixed & 28.87 & 29.59 & 29.63 & 30.12 & 29.25 & 30.15 & \textbf{31.52} & 27.87 & 28.35 & 27.47 & 28.34 & 26.95 & 27.66 & \textbf{29.46}\tabularnewline
\hline 
\multirow{5}{*}{SSIM$\text{\ensuremath{\uparrow}}$} & 10\textasciitilde 20\% & 0.949 & 0.960 & 0.955 & 0.956 & 0.957 & 0.968 & \textbf{0.969} & 0.941 & 0.947 & 0.936 & 0.943 & 0.939 & 0.946 & \textbf{0.951}\tabularnewline
 & 20\textasciitilde 30\% & 0.912 & 0.932 & 0.924 & 0.931 & 0.928 & 0.944 & \textbf{0.946} & 0.894 & 0.909 & 0.890 & 0.909 & 0.895 & 0.903 & \textbf{0.911}\tabularnewline
 & 30\textasciitilde 40\% & 0.872 & 0.901 & 0.889 & 0.902 & 0.895 & 0.916 & \textbf{0.917} & 0.844 & 0.867 & 0.841 & \textbf{0.872} & 0.848 & 0.852 & 0.865\tabularnewline
 & 40\textasciitilde 50\% & 0.830 & 0.868 & 0.853 & 0.871 & 0.862 & 0.883 & \textbf{0.885} & 0.790 & 0.824 & 0.789 & \textbf{0.831} & 0.802 & 0.775 & 0.815\tabularnewline
 & Mixed & 0.929 & 0.944 & 0.945 & 0.948 & 0.941 & 0.951 & \textbf{0.960} & 0.923 & 0.927 & 0.921 & 0.935 & 0.918 & 0.920 & \textbf{0.938}\tabularnewline
\hline 
\multirow{5}{*}{FID$\downarrow$} & 10\textasciitilde 20\% & 5.92 & 3.78 & 5.13 & 4.72 & 3.45 & 3.23 & \textbf{3.13} & 10.78 & 10.08 & 13.49 & 12.73 & 9.16 & 9.09 & \textbf{8.22}\tabularnewline
 & 20\textasciitilde 30\% & 10.31 & 6.33 & 8.78 & 7.06 & 5.36 & 5.29 & \textbf{4.87} & 18.90 & 17.39 & 24.67 & 20.46 & 15.96 & 16.32 & \textbf{14.32}\tabularnewline
 & 30\textasciitilde 40\% & 15.67 & 9.03 & 13.21 & 9.56 & 7.31 & 7.61 & \textbf{6.92} & 26.42 & 25.63 & 37.86 & 28.02 & 21.95 & 24.15 & \textbf{20.17}\tabularnewline
 & 40\textasciitilde 50\% & 21.92 & 12.19 & 18.86 & 12.58 & 9.15 & 9.86 & \textbf{8.85} & 35.61 & 34.82 & 54.45 & 35.75 & 27.22 & 37.93 & \textbf{25.68}\tabularnewline
 & Mixed & 8.28 & 5.38 & 5.27 & 5.20 & 4.57 & 4.54 & \textbf{3.78} & 15.05 & 13.68 & 15.74 & 15.04 & 13.05 & 13.41 & \textbf{10.42}\tabularnewline
\hline 
\end{tabular}
\end{table*}

We use PSNR, SSIM~\cite{wang2004image}, and Frechet Inception Distance (FID)~\cite{heusel2017gans} to evaluate the quantitative results on Celeba-HQ and Places2 in Table~\ref{table_quantitative}. As other inpainting researches, we randomly generate irregular masks~\cite{yu2019free} with $(10\%, 20\%], (20\%, 30\%], (30\%, 40\%], (40\%, 50\%]$ mask ratios. Besides, the mixed masks of both irregular mask and segmentation mask are tested to verify the effectiveness of real-word applications~\cite{zeng2020high}, \emph{e.g.}, object removal.
Our method achieves the best PSNR and FID in both Celeba-HQ and Places2 of all mask types. Although EC and MTS can get slightly better SSIM in Places2, they are specialized for the structural repair. Moreover, MTS needs much more parameters and computation due to its multi-layer attention compared with our method (Table~\ref{tab:eff_param}).
Note that HiFill fails to get satisfactory results as the mask areas are enlarged. We infer that the feature aggregation is not good for large regions missing. And a better feature encoding for masked regions is necessary, which is solved by AT in our method.
Co-Mod also achieves good results, but it is trained in larger data scales with a much more complex StyleGAN2 model~\cite{karras2020analyzing}. 
Note that our method works better than the state-of-the-art WaveFill, which also leverages the wavelet information during the inpainting.
In particular, our method achieves substantial advantages in FID compared with other methods, which indicates that our results are closer to human perception.


\subsection{Qualitative Comparisons}

\begin{figure*}
\begin{centering}
\begin{tabular}{c}
\includegraphics[width=0.95\linewidth]{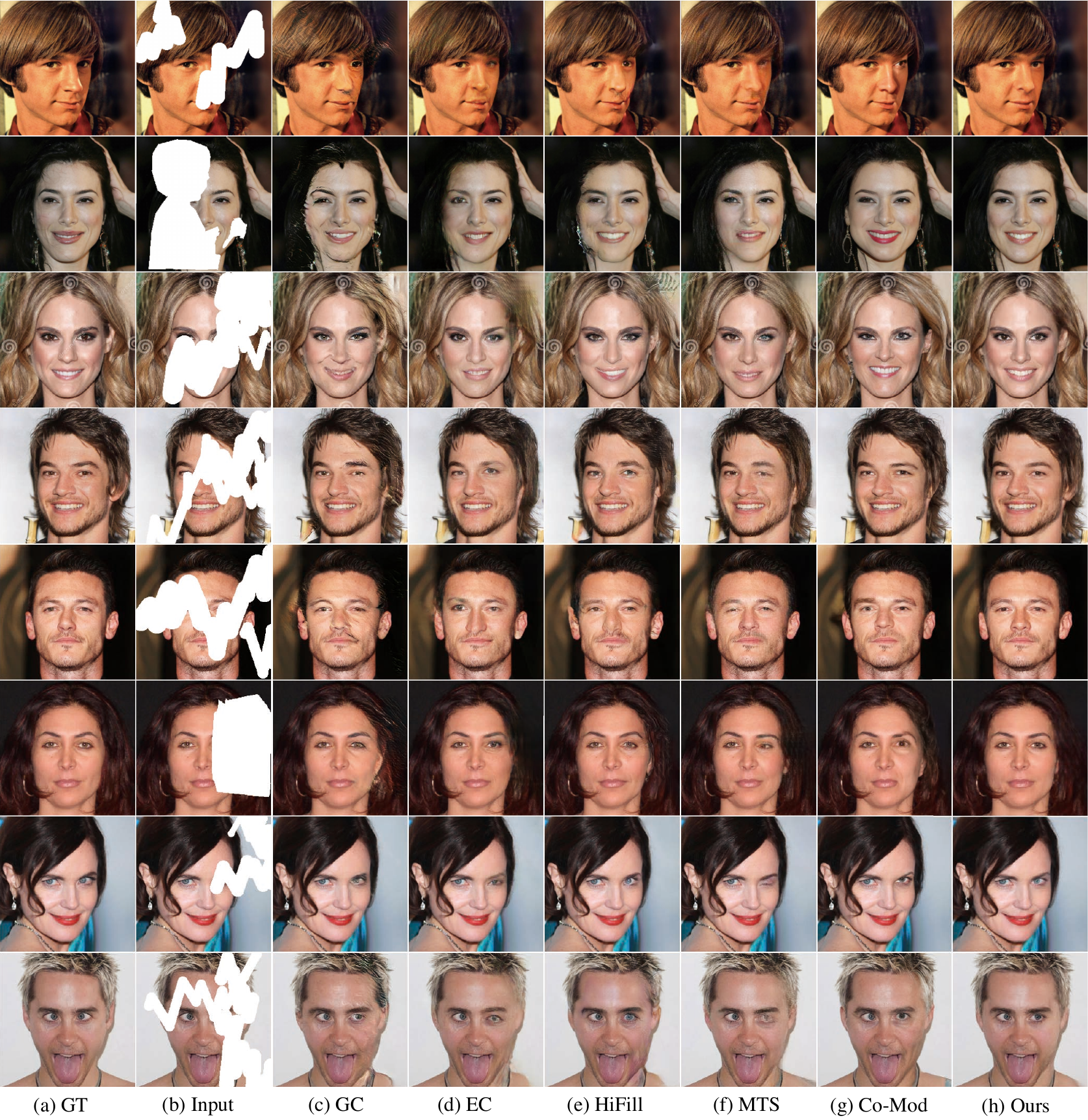} \tabularnewline
\end{tabular}
\vspace{-0.1in}
\caption{Qualitative results in Celeba-HQ with mixed masking. From left to right, ground truth (GT), masked inputs, Gate Convolution (GC)~\protect\cite{yu2019free}, Edge Connect (EC)~\protect\cite{nazeri2019edgeconnect}, Contextual Residual Aggregation for High-Resolution inpainting (HiFill)~\protect\cite{yi2020contextual}, Multi-Task Structural learning (MTS)~\protect\cite{yang2020learning}, Co-Modulation GAN (Co-Mod)~\protect\cite{zhao2021comodgan}, and ours inpainting results. Best viewed with zoom-in.}
\vspace{-0.1in}
\label{figure_celebahq_qualitative}
\end{centering}
\end{figure*}

\begin{figure*}
\begin{centering}
\begin{tabular}{c}
\includegraphics[width=0.95\linewidth]{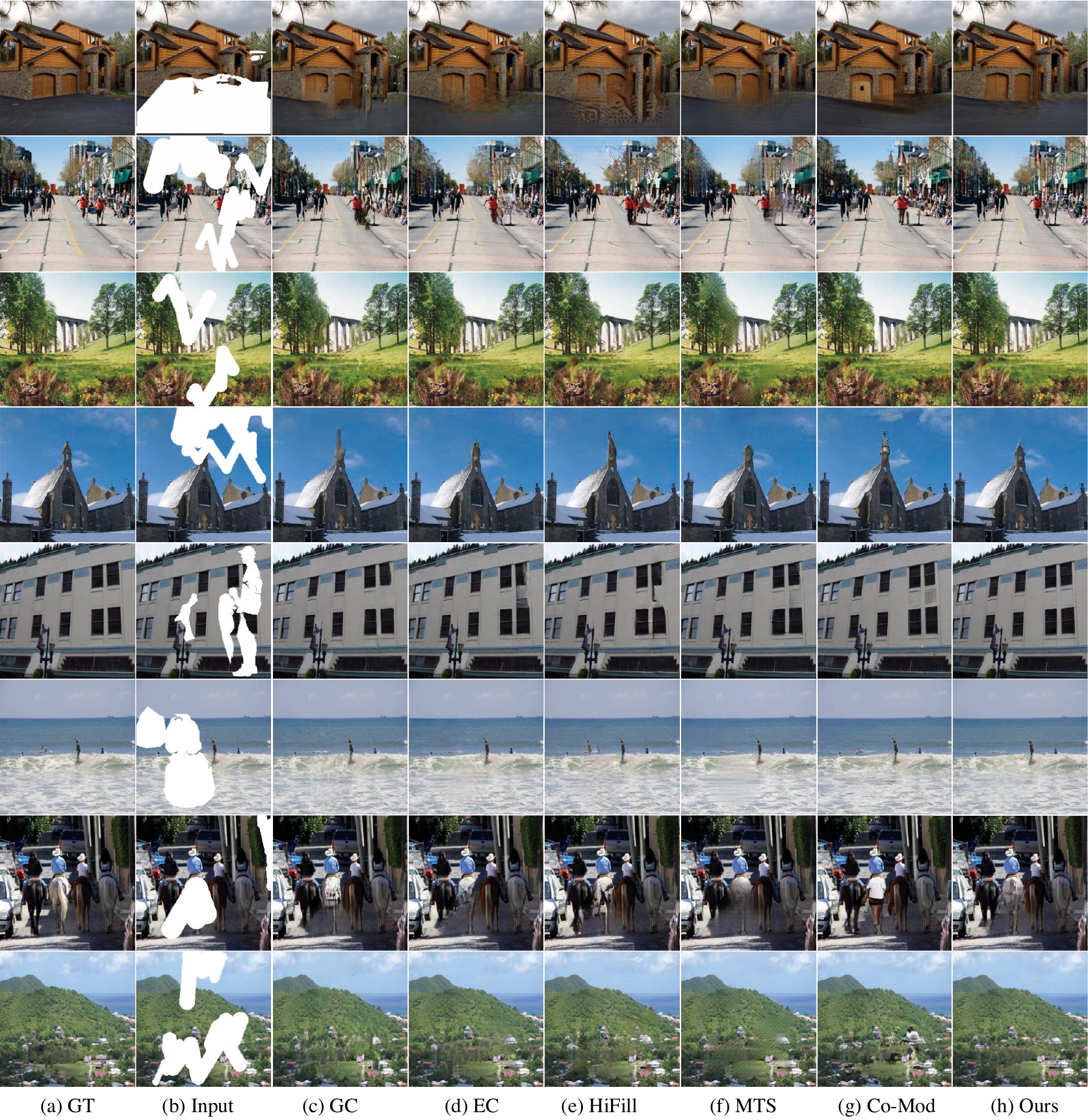} \tabularnewline
\end{tabular}
\vspace{-0.1in}
\caption{Qualitative results in Places2 with mixed masking. From left to right, ground truth (GT), masked inputs, Gate Convolution (GC)~\protect\cite{yu2019free}, Edge Connect (EC)~\protect\cite{nazeri2019edgeconnect}, Contextual Residual Aggregation for High-Resolution inpainting (HiFill)~\protect\cite{yi2020contextual}, Multi-Task Structural learning (MTS)~\protect\cite{yang2020learning}, Co-Modulation GAN (Co-Mod)~\protect\cite{zhao2021comodgan}, and ours inpainting results. Best viewed with zoom-in.}
\vspace{-0.1in}
\label{figure_places2_qualitative}
\end{centering}
\end{figure*}

In Figure \ref{figure_celebahq_qualitative} and Figure \ref{figure_places2_qualitative}, we show qualitative results of the proposed method compared with other state-of-the-art approaches on Celeba-HQ and Places2 respectively. In general, our method can achieve more reliable results. Especially, for the human face images in Celeba-HQ, the proposed method can reconstruct the face edge seamlessly and naturally, while EC, MTS suffers from blurring effect, and GC, HiFill have significant artifacts in the face boundaries. Therefore, the proposed wavelet prior attention can capture superior frequency correlations and further determine precise structural borderlines. Besides, although Co-Mod can achieve good images, it suffers from asymmetrical results, and loses high fidelity. Specifically, eye directions (rows 1) and facial structure (row 2) generated by Co-Mod are unnatural compared with our method as shown in Figure \ref{figure_celebahq_qualitative}.
In Places2, GC fails to recover complex details without artifacts. And EC works better in structures with edge prior, but it suffers from unreliable texture consistency. HiFill enjoys a little better texture in waves but it suffers from obvious artifacts in other instances. Besides, MTS generates blurring textures in natural and man-made scenes of Places2. Benefited from the complete GAN loss and large-scale training data, Co-Mod can achieve comparable results with ours, but it suffers from some uncontrollable generations, \emph{e.g.}, the abrupt door with a hole and inconsistent color (row 1) in Figure \ref{figure_places2_qualitative}. Whereas, our method can achieve both texture consistency and reasonable structures in the natural scenes.

\subsection{Ablation Studies and Discussions}
\label{sec:ablation}


\begin{table}
\centering
\vspace{-0.1in}
\caption{Ablation studies with different attention mechanisms and settings on Celeba-HQ.}
\vspace{-0.1in}
\label{table_ablation}
\renewcommand\tabcolsep{3.5pt}
\begin{tabular}{cccc|ccc}
\toprule
Datasets & \multicolumn{3}{c|}{Celeba-HQ} & \multicolumn{3}{c}{Places2}\tabularnewline
\midrule
Metrics & PSNR$\uparrow$ & SSIM$\uparrow$ & FID$\downarrow$ & PSNR$\uparrow$ & SSIM$\uparrow$ & FID$\downarrow$\tabularnewline
\midrule
CosAttention (Baseline) & 30.35 & 0.948 & 4.46 & - & - & -\tabularnewline
Standard Attention & 30.14 & 0.947 & 4.61 & - & - & -\tabularnewline
Baseline (512) & 30.47 & 0.949 & 4.33 & - & - & -\tabularnewline
Baseline+FA & 30.24 & 0.947 & 4.49 & - & - & -\tabularnewline
\midrule
Baseline+WPA & 31.05 & 0.956 & 4.09 & 29.14 & 0.934 & 11.27\tabularnewline
Baseline+WPA+ATs & \textbf{31.52} & \textbf{0.960} & \textbf{3.78} & \textbf{29.46} & \textbf{0.938} & \textbf{10.42}
\tabularnewline
\bottomrule
\end{tabular}
\end{table}

\begin{table}
\centering
\vspace{-0.1in}
\caption{Ablation studies of EMD$\downarrow$ with different frequencies (image sizes) on Celeba-HQ.}
\vspace{-0.1in}
\label{tab:emd_compare}
\begin{tabular}{cccccc}
\toprule
WPA & ATs & HF(256) & HF(128) & LF(64) & LF(32)\tabularnewline
\midrule
$\times$ & $\times$ & 0.2384 & 0.2384 & 0.2497 & 0.3080\tabularnewline
$\checkmark$ & $\times$ & 0.2297 & 0.2260 & 0.2439 & 0.2953\tabularnewline
$\times$ & $\checkmark$ & 0.2263 & 0.2243 & 0.2433 & 0.3000\tabularnewline
$\checkmark$ & $\checkmark$ & \textbf{0.2220} & \textbf{0.2189} & \textbf{0.2387} & \textbf{0.2852}\tabularnewline
\bottomrule
\end{tabular}
\end{table}


\begin{figure}
\begin{centering}
\includegraphics[width=1.0\linewidth]{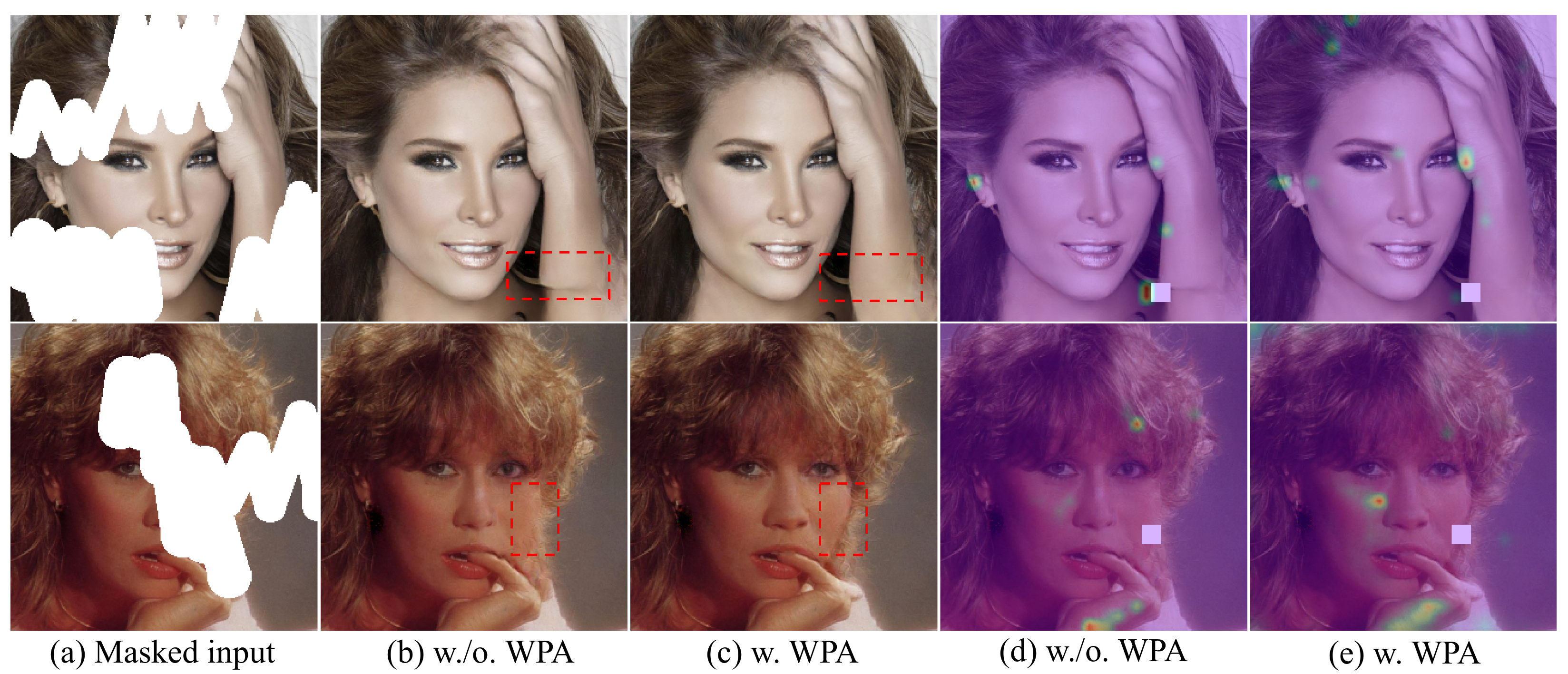} 
\par\end{centering}
\vspace{-0.1in}
\caption{Qualitative ablation results of WPA in Celeba-HQ. (a) masked input images; (b), (c) indicate the inpainted results of the inpainting model without/with WPA respectively. The main different areas are framed by red dotted boxes. Moreover, (d), (e) show the visualizations of attention maps (heatmaps) for the key regions (white patches) without/with WPA. Best viewed with zoom-in.}
\vspace{-0.1in}
\label{figure_qualitative_ablation}
\end{figure}

\begin{figure}[htbp]
\centering
\subfigure[FID]{
\includegraphics[width=4cm]{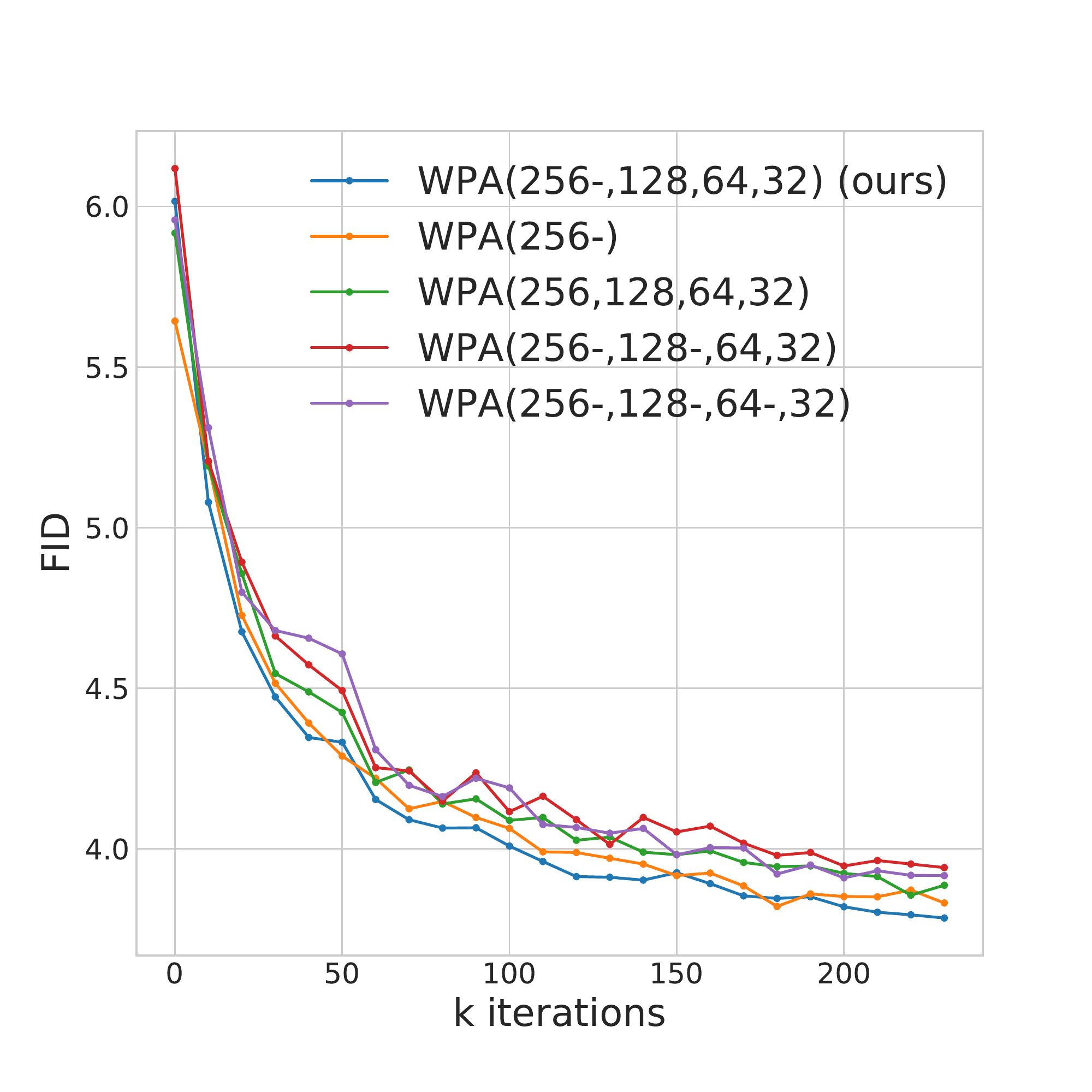}
}
\subfigure[PSNR]{
\includegraphics[width=4cm]{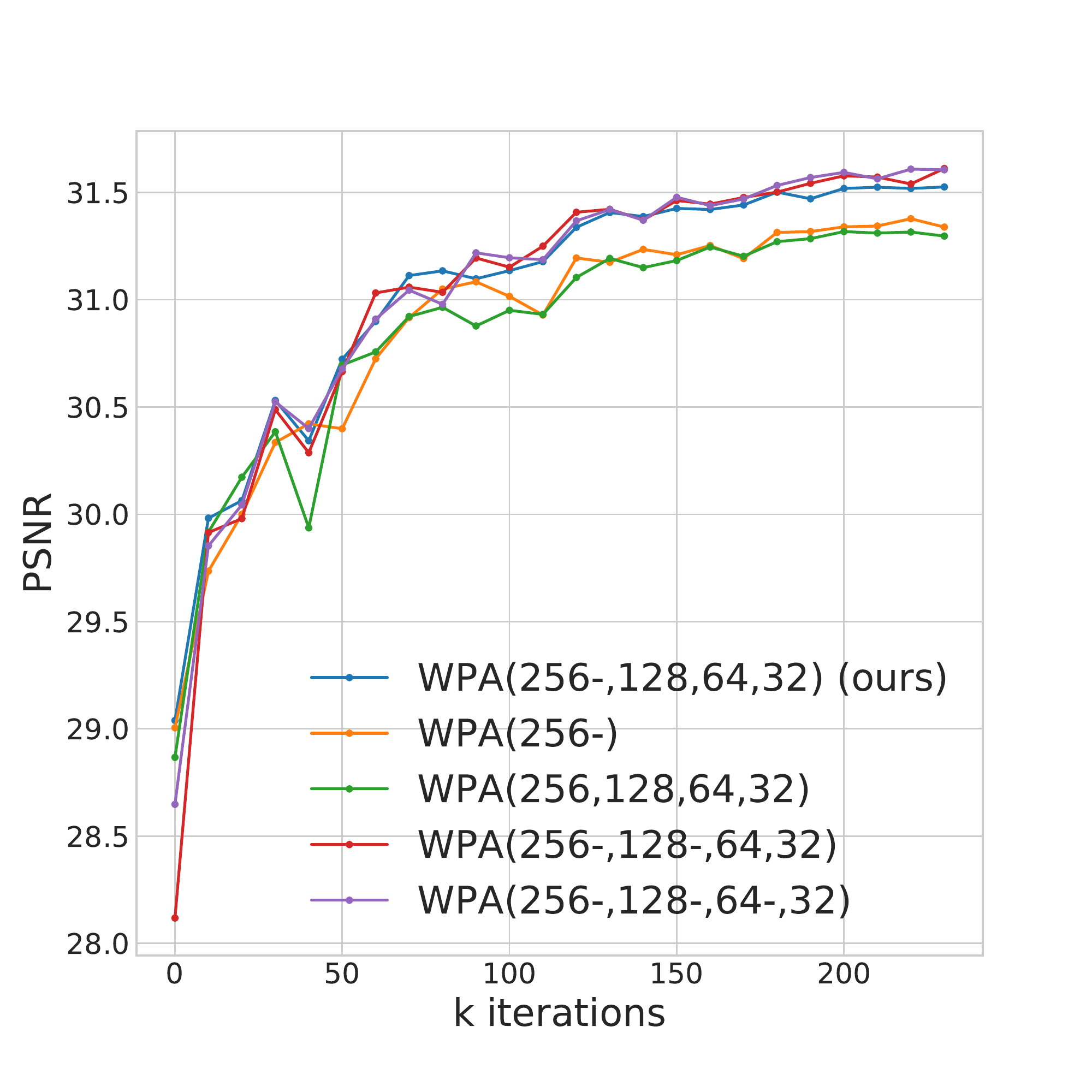}
}
\vspace{-0.1in}
\caption{Ablation studies on (a) FID, (b) PSNR of our method with different WPA settings. Values in parentheses indicate the wavelet sizes used in the comparison. `-' followed by the wavelet size means that high-frequency data is filtered by resizing for this scale.}
\vspace{-0.1in}
\label{fig:wpa_quantitative}
\end{figure}

\begin{figure}[htbp]
\centering
\subfigure[FID]{
\includegraphics[width=4cm]{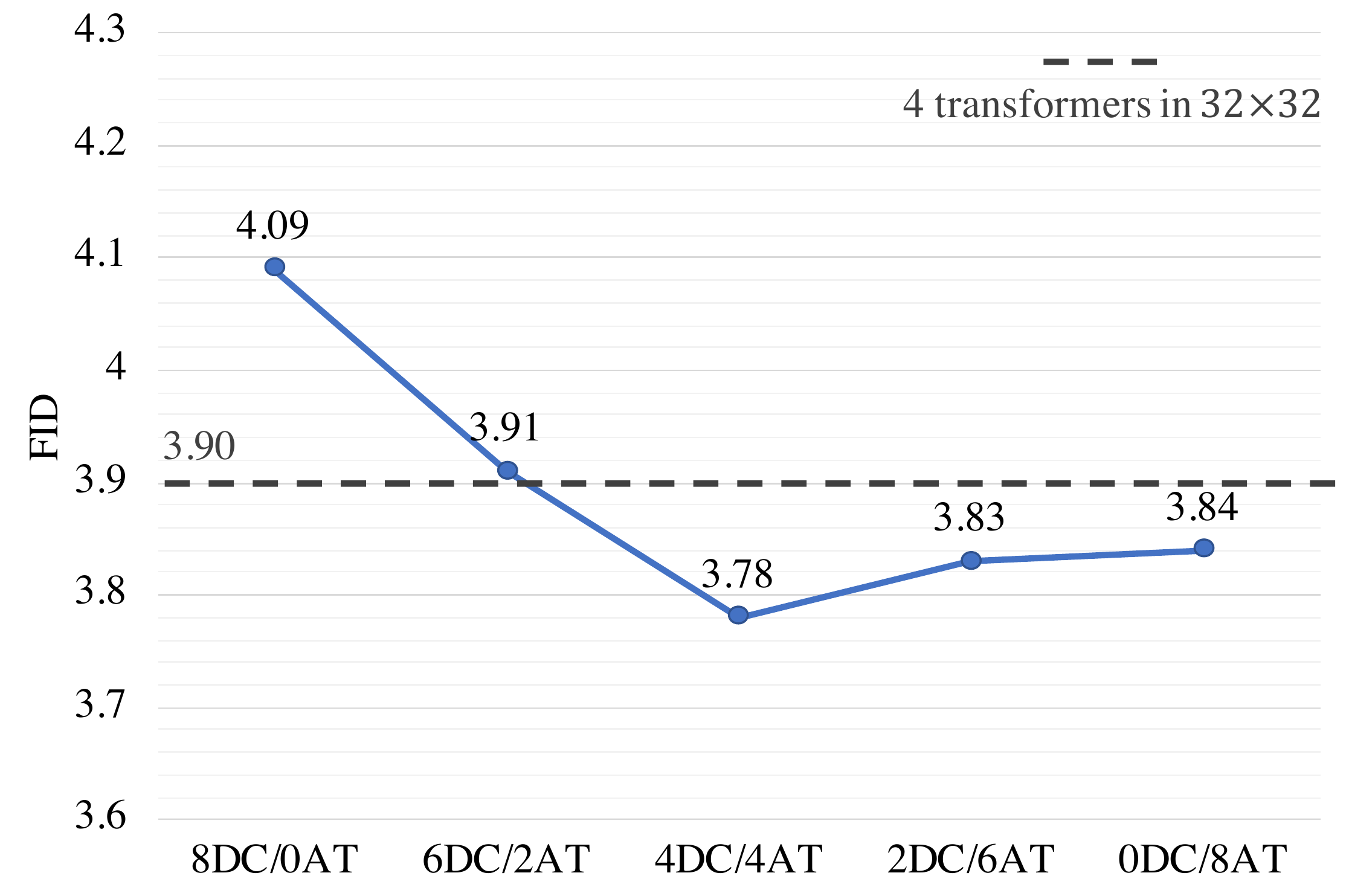}
}
\subfigure[PSNR]{
\includegraphics[width=4cm]{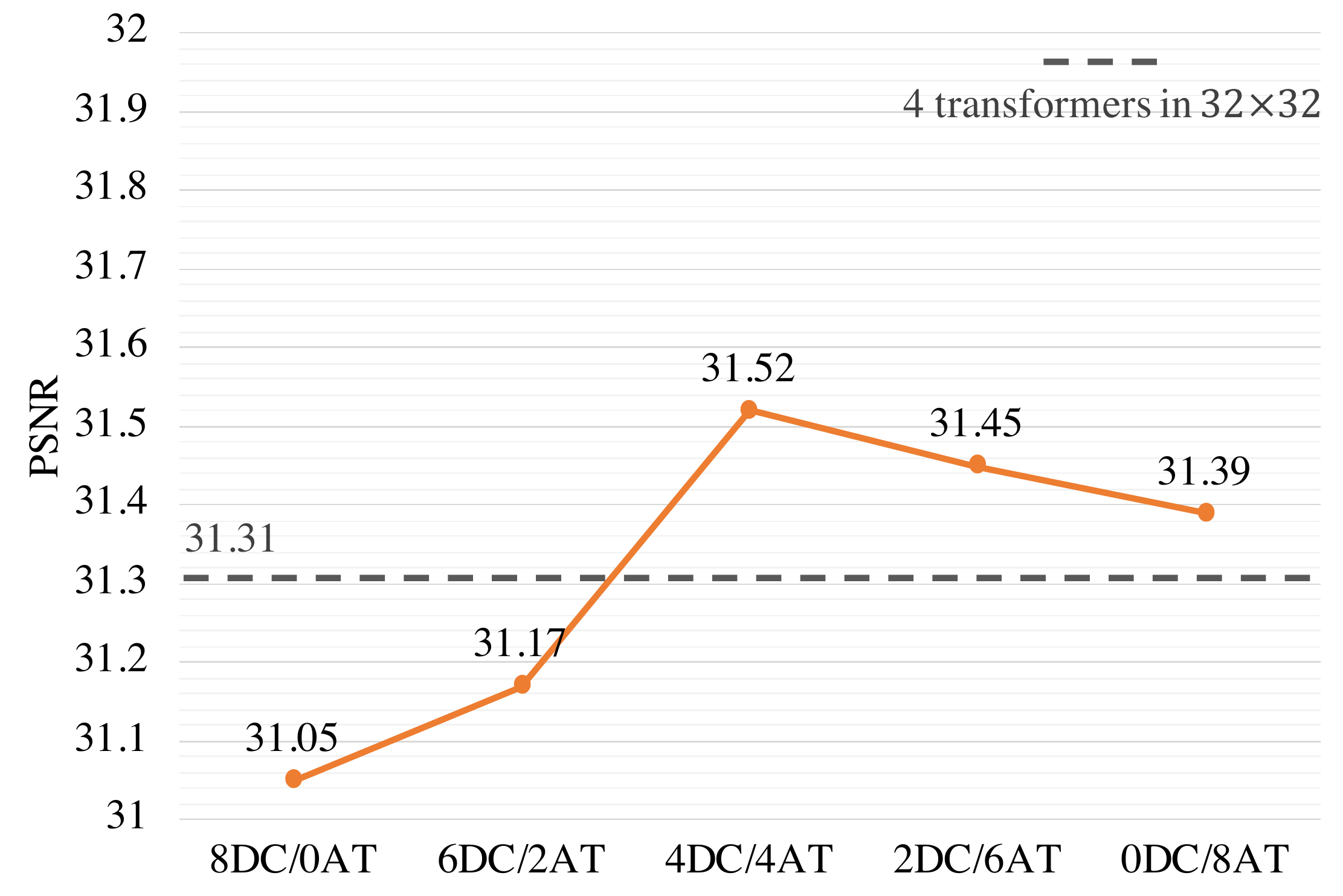}
}
\vspace{-0.1in}
\caption{Ablation studies on (a) FID, (b) PSNR of our model with different numbers of dilated convolutions (DCs) \& axial-transformers (ATs). The dotted line indicate the result of our model with 4 standard transformer blocks in $32\times32$.}
\vspace{-0.1in}
\label{fig:AT_quantitative}
\end{figure}

 \subsubsection{Quantitative Ablation of Different Attention Mechanisms}
 \label{Quantitative Ablation of Different Attention Mechanisms}
 
Quantitative ablation studies of different attention mechanisms are shown in Table~\ref{table_ablation}. The performance of standard self-attention~\cite{vaswani2017attention} falls behind the baseline model with CA~\cite{yu2018generative}, which indicates that CA with normalization is more suitable to the image inpainting. For simplicity, we take the CA based generator without WPA and ATs as the `baseline'. 
We also compare the baseline model with $512\times512$ inputs called baseline (512) to explore whether high-resolution information can benefit the network.
Although 512 inputs achieve slightly better results, our WPA still has considerable advantages compared with it. 
Furthermore, we compared the WPA with skip connection of feature-based attention aggregation (FA) used in HiFill~\cite{yi2020contextual}, which aggregates encoded features with different scales, and then skip-connects them to the related features in the decoder blocks. Note that this schema suffers from expensive computing costs. However, FA fails to improve the inpainting performance in our experiments. Note that ATs can further improve impressive performance in both Celeba-HQ and Places2 based on WPA, which indicates that the ATs are orthogonal to the WPA.

 \subsubsection{Quantitative Ablation of WPA} 
 \label{Quantitative Ablation of WPA}

To explore the influence of wavelets with different forms, we show the quantitative line charts of FID and PSNR in Figure~\ref{fig:wpa_quantitative}. Wavelets with smaller sizes (128,64,32) slightly improve the FID.
As discussed in Sec.~\ref{WPA}, we filter some high-frequency details by the resizing technique, which is shown as `-' followed by the wavelet size in Figure~\ref{fig:wpa_quantitative}. Interestingly, WPA with complete high-frequency information (256,128,64,32) works worse than the one with filtered $256\times256$ wavelet (256-,128,64,32). Two reasons are speculated as 1) since the ground truth size is still 256, high-frequency information from 512 size is unnecessary and hurts the metrics; 2) excessive high-frequency information is negative to the attention learning. We further try to filter wavelets of other sizes, which leads to worse FID scores. 

Besides, Earth Mover's Distances (EMD)~\cite{rubner2000earth} are compared in Table~\ref{tab:emd_compare} with and without WPA and AT based on the baseline model. In detail, we convert inpainted images into HSV space firstly. Then, normalized histograms are calculated in the masked regions. And EMD is achieved to measure the distribution distance from the ground truth HSV histogram. The comparisons are analyzed among different frequencies, \emph{i.e.}, image sizes. Although AT can also improve the EMD, the improvements are not prominent in low-frequency images \emph{e.g}, $64\times64$ and $32\times32$. The proposed WPA can achieve significant improvements in all sizes.

\subsubsection{Qualitative Ablation of WPA} 
\label{Qualitative Ablation Study of WPA}
From Figure~\ref{figure_qualitative_ablation}, We show some qualitative results between the baseline model with CA and our WPA. Benefiting from correctly distinguishing textures with different frequencies, our model can get more reliable textures and structural borderlines in the red dotted boxes. Furthermore, visualizations of the attention map in Figure \ref{figure_qualitative_ablation} (d) and (e) show that our model can pay correct attention to the missing regions. In the first row, our model pays more attention to the hand while the baseline attends to the hair nearby. And in the second row, our model can focus on the face rather than the hair, but the baseline pays more attention to the hair which causes blur in the face border. Substantially, our WPA can offer better receptive fields according to different-frequency information. It can further achieve both reliable texture and nearby structural clues according to the wavelet prior.

\subsubsection{Quantitative Ablation of ATs} 
\label{Quantitative Ablation Study of AT} 

In Figure~\ref{fig:AT_quantitative}, we show line charts of FID and PSNR of our model with different numbers of dilated convolutions (DCs) and ATs. Specifically, DCs are gradually replaced with ATs from the middle to both ends of the generator. The performance is improved along with the increasing AT blocks at first. The best result is achieved where AT\&DC is 4\&4 in our experiments. It indicates that DC and AT are complementary. DC pays more attention to local features, while AT is devoted to the feature dependencies with arbitrary distances for the global feature encoding. Besides, since the heavy computation of standard transformer (ST), we can only train the model with 4 STs and 4 DCs in $32\times32$ size. And the feature channels are doubled for ST to balance the capacity. The performances of the ST\&DC combination are shown as dashed lines in Figure~\ref{fig:AT_quantitative}, which works worse than our AT\&DC.

\subsubsection{Qualitative Ablation of ATs}
\label{Qualitative Ablation Study of AT} 

\begin{figure*}
\begin{centering}
\includegraphics[width=0.95\linewidth]{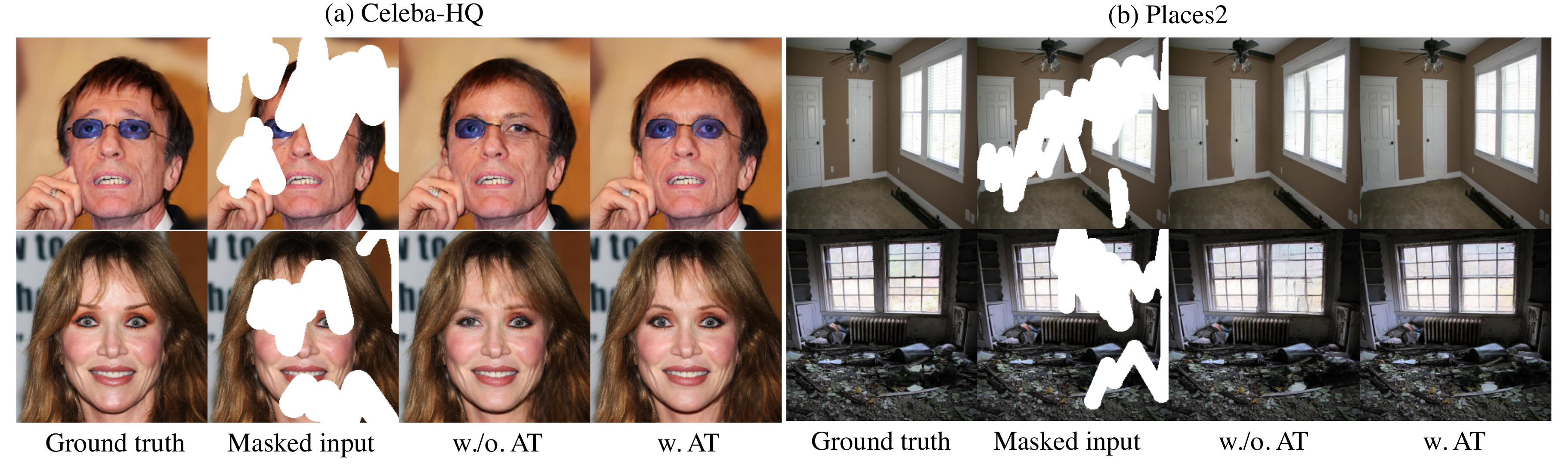} 
\par\end{centering}
\vspace{-0.1in}
\caption{Qualitative ablations of ATs on (a) Celeba-HQ and (b) Places2. More attentions should be paid to the recovery of eyes, glasses and structural edges. Please zoom-in for details.}
\vspace{-0.1in}
\label{fig:at_qualitative}
\end{figure*}

We show the qualitative ablation study of AT in Figure~\ref{fig:at_qualitative}. Note that the model without AT also contains CA to tackle the attention mechanism. For instances of Celeba-HQ, if one eye is blocked with the mask, the generator without AT fails to recover reasonable and symmetrical eyes or glasses, even with the attention module. We consider that the CA mechanism only tries to transfer textures from unmasked regions to the masked ones without long-range semantic restructuring. For the indoor environments in Places2, the corrupted image remains some unmasked clues for the structures, \emph{e.g.}, door corners, and windows. Our model with AT can recover much more precise structures with these clues compared with the one without AT.

\subsubsection{Time Efficiency and Model Parameters}

\begin{table}
\begin{center}
\renewcommand\tabcolsep{4.5pt}
\caption{Efficiency and parameter-scale ablation studies.}
\label{tab:eff_param}
\vspace{-0.1in}
\begin{tabular}{cccc}
\toprule
Methods & img/s(train)$\uparrow$ & img/s(test)$\uparrow$ & params\tabularnewline
\midrule
Baseline & 29.06 & 91.65 & 13.87M\tabularnewline
Baseline (512) & 26.93 & 78.75 & 13.92M\tabularnewline
Baseline+FA & 17.22 & 71.58 & 15.19M\tabularnewline
HiFill & 17.60 & 48.52 & 10.19M\tabularnewline
MTS & 11.48 &  19.18  & 17.85M \tabularnewline
Baseline+WPA (ours) & 26.67 & 87.69 & 14.06M\tabularnewline
\midrule
Baseline+WPA+ST*4 (64) & 8.44 & 26.92 & 12.53M\tabularnewline
Baseline+WPA+ST*4 (32) & 21.62 & 65.78 & 31.49M\tabularnewline
Baseline+WPA+AT*4 (64) & 20.51 & 63.05 & 13.59M\tabularnewline
\bottomrule
\end{tabular}
\end{center}
\vspace{-0.1in}
\end{table}

We test the model efficiency based on a single V100 GPU in Table~\ref{tab:eff_param}. The baseline model is our method without WPA and AT. The top half of Table~\ref{tab:eff_param} shows efficiency without transformer blocks for the generator. 
Our WPA is more efficient than larger inputs, feature-based attention aggregation (FA).
For the bottom half of Table~\ref{tab:eff_param}, we show the results of standard transformers (STs) and Axial-transformers (ATs). Replacing DCs by STs reduces the parameters because ST is consisted of linear layers with fewer parameters compared with convolutional ones. STs in $64\times64$ are extremely costly to train with terrible efficiency. Downsampling the middle features into $32\times32$ with doubled channels can reduce the computation. Our $64\times64$ ATs outperform $32\times32$ STs in Figure~\ref{fig:at_qualitative} with almost the same speed.

\subsection{Comparison with WaveFill}
\begin{figure*}
\begin{centering}
\includegraphics[width=1\linewidth]{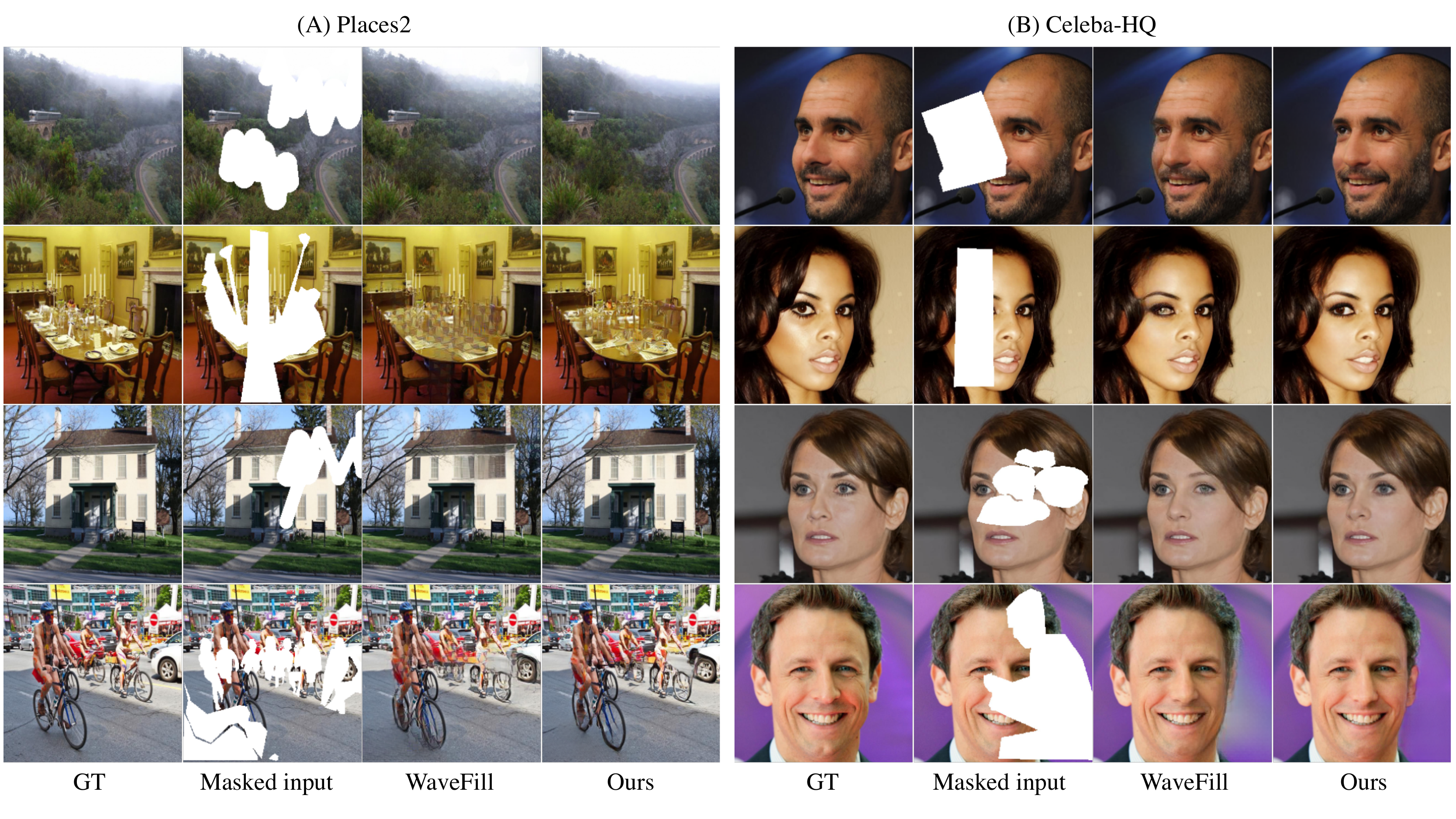} 
\par\end{centering}
\caption{Qualitative comparisons between our method and WaveFill~\protect\cite{yu2021wavefill} on (A) Places2 and (B) Celeba-HQ. For each sub figure, ground truth (GT), masked inputs, results of WaveFill, results of ours are shown from left to right.}
\label{fig:wavefill_qualitative} 
\end{figure*}

As a state-of-the-art inpainting method which also leverages the wavelet, we further compare some qualitative results with WaveFill~\cite{yu2021wavefill} in Fig.~\ref{fig:wavefill_qualitative}. WaveFill can get good results in face images, but it can not tackle complex scene situations in Places2. Moreover, for the face images from Celeba-HQ, our method still achieves better recovery of the eye consistence and facial boundaries. Note that our method can be trained more efficiently compared with WaveFill, and outperform it with complementary Wavelet image Prior Attention (WPA) and stacked Axial-Transformer blocks (ATs).

\subsection{User study}

\begin{figure}
\begin{centering}
\includegraphics[width=0.95\linewidth]{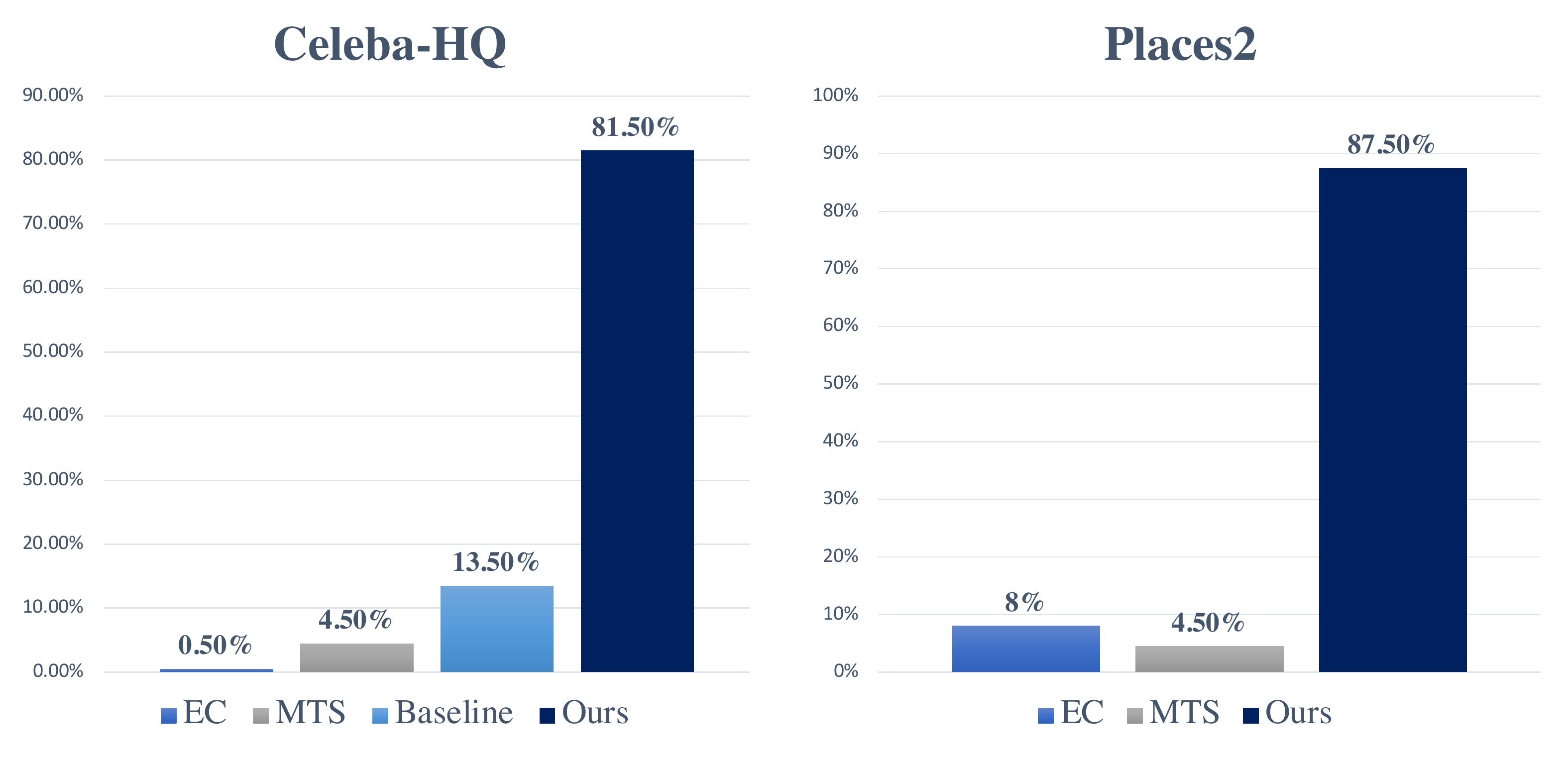} 
\par\end{centering}
\caption{Average scores of Celeba-HQ and Places2 for user studies, which are collected from volunteers who select the best one from shuffled results.}
\label{fig:user_study}
\end{figure}

Quantitative metrics fail to reflect the objective quality of inpainted images. Therefore, we provide user studies to ensure the effectiveness of our proposed method as shown in Figure~\ref{fig:user_study}. Specifically, we invite 10 volunteers who are not familiar with computer vision and image inpainting to judge the quality of inpainted results. For Places2 we compare 3 methods which include EC~\cite{nazeri2019edgeconnect}, MTS~\cite{yang2020learning}, and ours. For Celeba-HQ, Baseline (GC+CA+DCs) is also participated in the comparison. Volunteers need to choose the best result generated by randomly shuffled competitors. As shown in Figure~\ref{fig:user_study}, our method can achieve the best result in human perception.


\section{Conclusion}

In this paper, we propose Wavelet prior attention learning in Axial Inpainting Network (WAIN), which extends the attention into both feature aggregation and feature encoding to improve the image inpainting. For the feature aggregation, a parameter efficient wavelet prior attention module, which leverages various frequency information to supervise the high-level attention map directly. The multi-scale strategy is also implemented for better generalization. Moreover, we leverage stacked Axial-Transformer to encode low-level masked features for meaningful semantics. Related image inpainting experiments prove the effectiveness and the efficiency of the proposed method.

\appendix

\section*{Network Architectures}

Details of some network architectures are illustrated as follows. 

\noindent\textbf{GateConv Block.} The GateConv is consist of GateConv2D~\cite{yu2019free} $\rightarrow$ InstanceNorm $\rightarrow$ ReLU. We also add the spectral normalization~\cite{miyato2018spectral} to the GateConv used in the generator.

\noindent\textbf{DilatedConv Block.}
The DilatedConv block is the same as the one used in EdgeConnect~\cite{nazeri2019edgeconnect} with Conv2D $\rightarrow$ InstanceNorm $\rightarrow$ ReLU $\rightarrow$ Conv2D $\rightarrow$ InstanceNorm, where the first convolution is based on $\mathrm{dilation}=2$.

\section*{Subset Selected from Places2}
As discussed in the main paper, we chose 10 categories randomly with 50k training images and 1k validation images from Places2~\cite{zhou2017places}, which include `valley', `church-outdoor', `village', `house', `dining room', `street', `ocean', `bow window-indoor', `boathouse', and `viaduct'. The selected subset of Places2 is ranged from urban to natural scenes, which is adequate to validate the effectiveness of inpainting methods.

\ifCLASSOPTIONcaptionsoff
  \newpage
\fi

\bibliographystyle{ieeetr}
\bibliography{IEEEabrv}

%

\begin{IEEEbiography}[{\includegraphics[width=1in,height=1.25in,clip,keepaspectratio]{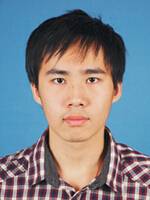}}]{Chenjie Cao} received the M.S. degree in Computer Science from East China University of Science and Technology, in 2019. He is currently pursuing the Ph.D. degree in Statistics from Fudan University. His research interests include machine learning, deep learning, and computer vision.
\end{IEEEbiography}

\begin{IEEEbiography}[{\includegraphics[width=1in,height=1.25in,clip,keepaspectratio]{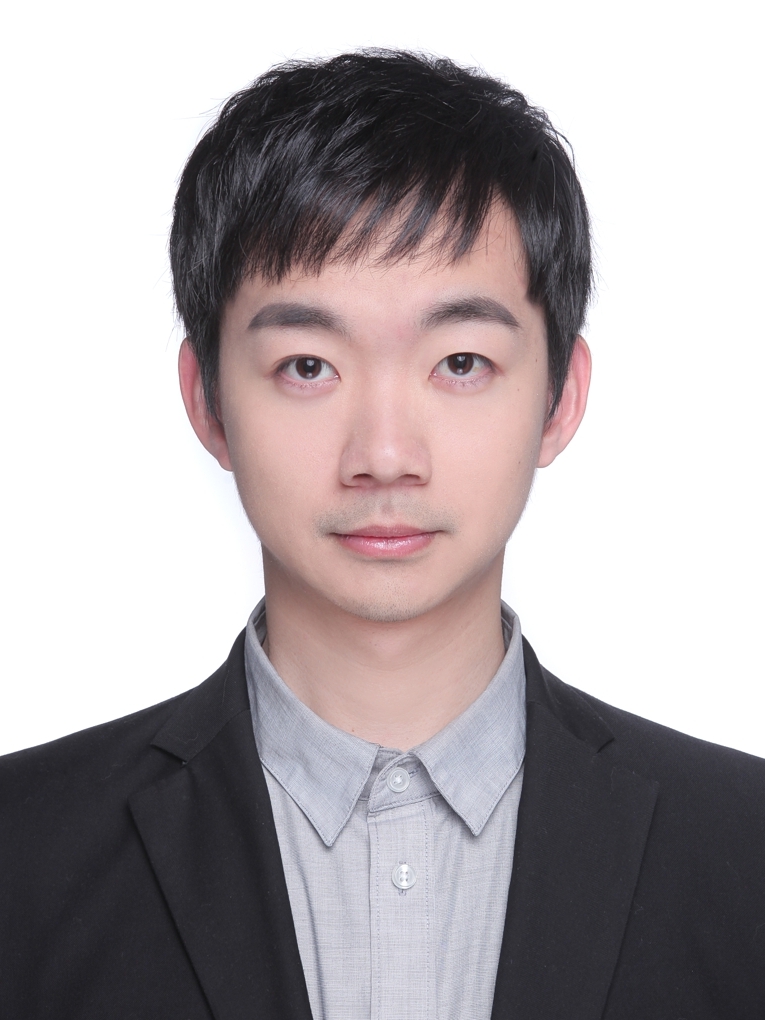}}]{Chengrong Wang}
received B.E. degree in Computer Science from ShanghaiTech, in 2019. He is currently pursuing a master's degree in software engineering at Fudan University. His research interests include machine learning, deep learning, and computer vision.
\end{IEEEbiography}

\begin{IEEEbiographynophoto}{Zhangyun Tao}
received B.E. degree in Statistics from Zhongnan University of Economics and Law, in 2020. She is currently pursuing a master's degree in the school of data science in Fudan University. Her research interests include machine learning, deep learning, and computer vision.
\end{IEEEbiographynophoto}

\begin{IEEEbiography}[{\includegraphics[width=1in,height=1.25in,clip,keepaspectratio]{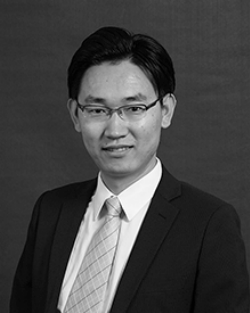}}]{Yanwei Fu} received his PhD degree from the Queen Mary University of London, in 2014. He worked as  post-doctoral research at Disney Research, Pittsburgh, PA, from 2015 to 2016. He is currently a tenure-track professor with Fudan University.  
He was appointed as the Professor of Special Appointment (Eastern Scholar) at Shanghai Institutions of Higher Learning.
He published more than 80 journal/conference papers including IEEE TPAMI, TMM, ECCV, and CVPR. His research interests are one-shot/meta learning, learning based 3D reconstruction, and image and video understanding in general. 
\end{IEEEbiography}

\end{document}